\documentclass[letterpaper,11pt,reqno]{amsart}

\makeatletter
\usepackage{amssymb}
\usepackage{latexsym}
\usepackage{amsbsy}
\usepackage{amsfonts}
\usepackage{comment}
\usepackage{hyperref}
\usepackage{graphicx}
\usepackage{etoc}
\usepackage{parskip}
\usepackage{scalerel}
\usepackage{enumerate}
\usepackage{enumitem}
\usepackage{mathtools}
\usepackage{color}
\usepackage{tcolorbox}
\usepackage{booktabs}
\usepackage{subcaption}
\usepackage{appendix}
\usepackage{algorithm}
\usepackage{algorithmic}
\usepackage{caption}
\usepackage{tikz}
\usepackage{tabularx}
\usetikzlibrary{arrows.meta, positioning}
\usepackage{fix-cm}

\def\marginpar#1{\ignorespaces}

\definecolor{codegreen}{RGB}{32,145,72}
\newcommand{\greencode}[1]{\textcolor{black}{\texttt{\detokenize{#1}}}}

\usepackage{natbib}
 \bibpunct[, ]{(}{)}{,}{a}{}{,}%
\numberwithin{equation}{section}
\theoremstyle{remark}

\makeatother
\begin{document}
\title[Spec2Twin-Chain]{Spec2Twin-Chain: Orchestrating Bi-Level Optimization with LLMs for Blockchain Digital Twin Construction}

\author[Haoting Zhang]{{Haoting} Zhang}
\address{University of California, Berkeley.}
\email{haoting\_zhang@berkeley.edu}

\author[Jiayuan Sheng]{{Jiayuan} Sheng}
\address{Columbia University.}
\email{js6646@columbia.edu}

\author[Haoxian Chen]{{Haoxian} Chen}
\address{Columbia University.}
\email{hc3136@columbia.edu}

\author[Donglin Zhan]{{Donglin} Zhan}
\address{Columbia University.}
\email{dz2478@columbia.edu}

\author[Zeyu Zheng]{{Zeyu} Zheng}
\address{University of California, Berkeley.}
\email{zyzheng@berkeley.edu}

\author[David D. Yao]{{David D.} Yao}
\address{Columbia University.}
\email{yao@columbia.edu}

\author[Wenpin Tang]{{Wenpin} Tang}
\address{Columbia University.}
\email{wt2319@columbia.edu}

\date{\today}
\begin{abstract}
Building a blockchain digital twin largely requires translating domain knowledge and specific system descriptions into a simulator architecture, calibrating its parameters against behavioral evidence, and validating the constructed twin. These steps are commonly performed through application-specific modeling efforts that can be difficult to reuse across systems and downstream decision problems. We consider automating this process through Spec2Twin-Chain, a framework that formulates blockchain digital-twin construction as a bi-level optimization problem. At the upper level, a large language model proposes and revises structurally admissible architectures using system specifications, behavioral evidence, and feedback from evaluated designs. At the lower level, a simulation-based optimizer calibrates the architecture-conditioned parameters under explicit objectives and guardrail constraints. The two levels iterate. The evaluated candidates at lower levels are retained in a global archive and used to guide subsequent proposals at upper levels. We conduct controlled experiments involving twin calibration, feedback-driven recovery, stress analysis, downstream policy optimization, and policy updating. The results demonstrate that the framework can construct behaviorally accurate twins, improve initial designs through iterative feedback, and reuse calibrated twins to support downstream decisions.
\end{abstract}
\maketitle



\section{Introduction}
\label{sec:introduction}

Blockchain systems now support applications beyond cryptocurrency, including finance, supply chain operations, healthcare, and IoT ecosystems.
These applications use distributed ledgers and smart contracts for record integrity, traceability, and automated execution. Their behavior depends on interactions among network conditions, incentive mechanisms, execution environments, and application-layer logic.

For instance, transaction-ordering dependencies in decentralized finance (DeFi) have given rise to adversarial strategies that translate application-layer dynamics into consensus-layer risks \citep{daian2019flash}. 
A notable example is the emergence of proposer-builder separation \citep{ethereum_pbs} as a new consensus mechanism, driven by MEV attacks.
Cross-chain bridges also introduce security risks that require careful evaluation before deployment \citep{belenkov2025sok}. These examples motivate tools for evaluating emergent behavior under realistic conditions instead of relying only on intuition or limited ad hoc testing.

Experimenting on live blockchains is often infeasible or unsafe. At the network layer, the scale and geographic distribution of nodes make faithful live evaluation prohibitively difficult \citep{aoki2019simblock}. At the application layer, the immutability of smart contracts and the real economic value at stake mean that ``testing in production'' can result in irreversible financial losses. While simulation is a standard alternative, ad hoc simulations rarely produce a validated representation that can be reused as operating conditions change.

Blockchain digital twins support monitoring and analysis, and let designers explore counterfactual scenarios in a controllable setting. In engineering, a digital twin links a physical system to a virtual representation. 
However, applying the digital twin concept to blockchains is more challenging, because blockchains emerge from tightly coupled network, consensus, and incentive mechanisms. Transactions and blocks propagate asynchronously, so nodes can temporarily hold different local views of pending transactions and the ledger \citep{decker2013information}, while consensus security and performance depend jointly on network and protocol parameters \citep{grevais2016}. A blockchain digital twin must therefore represent both system structure and architecture-conditioned operational parameters, consistent with blockchain simulation frameworks such as BlockSim \citep{alharby2020blocksim}. Adoption is hindered by the lack of universal reference frameworks, and by the effort required to maintain a twin as the protocol evolves \citep{sharma2022digital}. Recent work has used digital twins to manage the trade-off between decentralization, scalability, and security \citep{diamantopoulos2022digital}, and has examined Dynamic Data-Driven Application Systems (DDDAS) for blockchain management \citep{diamantopoulos2022dynamic}. Building a blockchain digital twin nevertheless requires substantial manual modeling and domain expertise.

\subsection*{Bespoke digital-twin construction}
Many blockchain modeling tools exist, yet constructing a digital twin remains largely a bespoke process. 
As indicated in \citep{albshri2022blockchain},
no single simulator encompasses the wide range of features required across diverse blockchain architectures. This fragmentation forces developers to make manual, often opaque modeling choices regarding abstraction levels, parameter calibration, and invariant validation. Furthermore, while the digital twin community emphasizes composability and reusability \citep{shao2020framework}, turning an informal system description into a validated twin still requires manual reconfiguration across toolchains, which slows iteration and introduces modeling errors \citep{zargham_rice_2019_cadcad}.

\subsection*{LLM-assisted digital-twin construction}
Recent progress in tool-using Large Language Models (LLMs) suggests treating digital-twin construction as a structured synthesis problem. LLM agents can interleave reasoning traces with executable actions, when interacting with external environments \citep{yao2023react}. In this setting, the LLM acts as an orchestrator rather than the simulator. 
It translates high-level goals into experiment specifications, and proposes modeling choices; after evaluation, it uses the results to revise the twin's configuration. The problem of interest is how an explicit design space and executable validation can support reliable LLM-guided synthesis.

\subsection*{Spec2Twin-Chain}
\begin{figure}
    \centering
    \includegraphics[width=\linewidth]{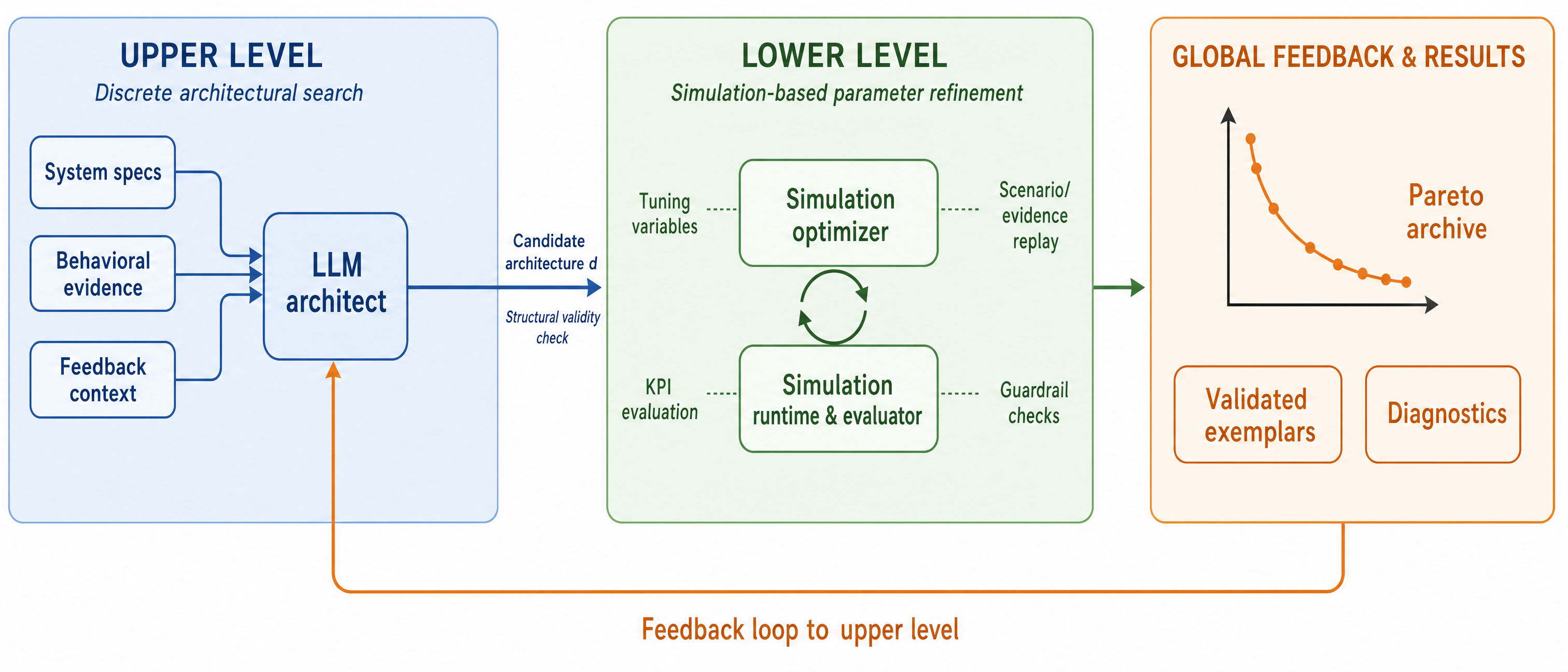}
    \caption{Bi-level optimization framework of \textit{Spec2Twin-Chain}. At the upper level, an LLM architect uses the system specification, behavioral evidence, and feedback context to propose a structurally valid candidate architecture $d$, including component choices and interaction logic. At the lower level, a simulation-based optimizer tunes the architecture-conditioned variables $\theta$, replays the target scenarios or evidence, evaluates KPI fit, stability, and cost objectives, and enforces guardrail checks. Feasible non-dominated configurations are accumulated in a global Pareto archive; validated exemplars and diagnostics are then fed back to the upper level to guide the next round of architectural search.}
    \label{fig:framework}
\end{figure}

Two complementary limitations exist in the conventional digital-twin constructions: bespoke processes require substantial effort on system-specific modeling, whereas LLM-assisted synthesis does not guarantee that a plausible proposal is admissible or survives executable validation.

To address the aforementioned challenges, we propose \textit{Spec2Twin-Chain}, a framework that facilitates blockchain digital-twin development using LLM agents. The workflow proceeds in three steps:
\begin{enumerate}
\item LLM agents translate the informal request and its behavioral evidence into the terms of a declarative schema, which specifies the admissible component architectures, parameter constraints, and safety invariants; every candidate twin is verified against this schema. 
This step allows us to transform blockchain digital-twin development into an optimization problem.

\item The builder then searches the admissible design space through bi-level optimization. At the upper level, an LLM agent proposes/sample candidate architectures concurred with the schema. At the lower level, for each valid proposal, a simulation-based optimizer tunes the variables that the architecture exposes.

\item The upper and lower levels iterate. The scores and diagnostics from each round of the lower level are recorded, and the feasible non-dominated designs form an archive. This archive supplements the context for the subsequent rounds of proposals at the upper level, so that the LLM agents learn from previous experience. Figure~\ref{fig:framework} summarizes this iterative loop.
\end{enumerate}

\section{Literature Review}
\label{sec:literature-review}

\subsection{Blockchain}
\label{sec:literature-blockchain}

Blockchain is a  distributed ledger,
where network participants use a consensus protocol to agree on valid transactions. Bitcoin provided an early and influential application of this design \citep{boehme2015bitcoin}, and the technology has since been applied to systems that require shared records, traceability, or programmable execution, including a broad range of IoT settings \citep{khan2024blockchain}. These applications also introduce new failure modes, particularly when assets and messages move across chains. Because live stress tests can expose networks and users to financial and operational risk, simulation provides a controlled setting for evaluating blockchain behavior under specified workloads and attacks.

\paragraph{Representative Applications.}
Blockchain has been studied in settings where several organizations need to maintain a shared record. In supply chain management, where a central concern is the tracking and coordination of goods from manufacturers to end customers \citep{tan2002supplychain}, such records provide a shared and auditable view across organizations. Examples include \emph{IBM Food Trust} \citep{IBM2018FoodTrust,Kawaguchi2019appofb} and the \emph{Everledger} platform \citep{CSIRO2022Everledger,Kshetri2022Block}. In healthcare, \emph{MedRec} uses blockchain-based permission management for access to medical records \citep{Azaria2016Contracts}. Energy applications use similar mechanisms for market coordination and auditable exchange \citep{teng2021review}; the \emph{TRANSAX} platform, for example, supports transactions within decentralized microgrids \citep{Laszka2018TRANSAX}. Decentralized finance provides another prominent application class: Uniswap implements decentralized trading through an automated market maker \citep{adams2021uniswapv3}, while Hyperliquid provides fully on-chain spot and perpetual-futures order books \citep{hyperliquid2026overview,hyperliquid2026orderbook}. Such applications incorporate protocol rules with transaction demand, liquidity, and ordering, with which we may simulate and study congestion and strategic interactions.

\paragraph{Cross-chain Interoperability and Security Incidents.}
Cross-chain protocols transfer assets or messages between otherwise separate networks. They are widely used in decentralized finance, but the bridge logic required for these transfers adds key-management, validation, and authorization risks. Broad surveys classify DeFi vulnerabilities and their associated attack patterns \citep{li2022defi}, and bridge-specific systematizations organize bridge components, exploit classes, and historical incidents \citep{lee2023bridge_sok,belenkov2025sok}. Automated tools such as \emph{Xscope} inspect cross-chain transactions for signs of exploitation \citep{zhang2023xscope}. Individual incident reports illustrate the recurring failure classes: fraudulent deposit-address registration after a key exposure \citep{Bifrost2022BifiBTCPostmortem}, privileged-invocation and verification bypasses \citep{Rekt2021PolyNetwork,Rekt2022Wormhole}, and missing recipient-contract validation \citep{QAN2022BridgeDisclosure}. Related key-management and authorization failures also occur outside bridge logic, as in the Ankr staking-token exploit \citep{Rekt2022AnkrHelio}.

\paragraph{Experimental Methodology and Benchmarks.}
Blockchain measurements depend on the workload, network configuration, and selected performance metrics, so experimental comparisons require an explicit benchmark design. \citet{grevais2016} use Markov decision processes to study double-spending and selfish-mining behavior in proof-of-work blockchains. For permissioned systems, \emph{BLOCKBENCH} combines end-to-end workloads with layer-specific microbenchmarks to identify bottlenecks across the blockchain software stack \citep{blockbench2017}. \emph{Gromit} instead uses a black-box transaction-fabric model and varies validator counts and network delays to compare scalability, throughput, latency, and sensitivity to network conditions across blockchain architectures \citep{Nasrulin2022gromit}.

\paragraph{Blockchain Simulations for Protocol Design.}

Simulation allows researchers to test protocol changes and adversarial conditions without deploying them on a live blockchain. Early studies modeled network-layer attacks against Bitcoin \citep{neudecker2015simulation}, and evaluated selfish-mining strategies \citep{gobel2016bitcoin}. \emph{BlockSim} provides configurable models of proof-of-work consensus and related blockchain processes \citep{alharby2020blocksim}. Other simulators focus on different protocol questions. \emph{BlockEmulator} provides modular support for deploying and testing blockchain sharding protocols and their cross-shard transaction mechanisms \citep{huang2025blockemulator}, while \emph{JABS} offers a modular environment for designing and evaluating consensus algorithms, including non-proof-of-work protocols \citep{jabs2023consensus}.

\subsection{LLM Agents}
\label{sec:literature-llm-agents}

LLM agents extend language models with mechanisms for planning, tool use, and memory. Surveys organize these systems around profile, memory, planning, and action components \citep{wang2024survey}, and planning-focused surveys further distinguish task decomposition, plan selection, external modules, reflection, and memory \citep{huang2024planning}. These capabilities are relevant to digital-twin construction because the model must translate a system description into executable artifacts, invoke simulation tools, and revise its proposal after receiving validation or performance results.

\paragraph{Tool Use and API Grounding.}
Tool use connects an LLM to simulators, data sources, and executable validation routines. \emph{ReAct} interleaves reasoning traces with external actions \citep{yao2023react}. \emph{Toolformer} and \emph{ToolLLM} study how models can learn or be instructed to invoke APIs \citep{schick2023toolformer,qin2023toolllm}. \emph{Gorilla} further shows that retrieving API documentation at inference time can reduce incorrect tool calls when interfaces change \citep{patil2023gorilla}. In our setting, these results support exposing simulators and blockchain RPC endpoints as tools with explicit input and output contracts.

\paragraph{Multi-Agent Decomposition.}
Multi-agent systems divide a workflow among model instances with different responsibilities. \emph{AutoGen} and \emph{CAMEL} use role definitions and structured exchanges to coordinate these instances \citep{wu2023autogen,li2023camel}. \emph{ChatDev} applies a similar design to software development by assigning separate roles to design, implementation, and testing \citep{qian2023chatdev}. Spec2Twin-Chain uses distinct worker roles for experiment design, execution, and verification so that each output can be checked before it enters the next stage.

\paragraph{Feedback, Memory, and Iteration.}
Iterative workflows also need a way to reuse earlier results without retraining the model. \emph{Reflexion} stores textual feedback in episodic memory for use in later attempts \citep{shinn2023reflexion}. \emph{Voyager} accumulates reusable programs and revises them after execution failures \citep{wang2023voyager}, while \emph{Generative Agents} organizes observations into a memory stream that supports later planning \citep{park2023generative}. These studies motivate retaining evaluated designs, objective values, and diagnostics as context for subsequent digital-twin proposals.

\paragraph{Interfaces and Guardrails.}
The interface between an agent and its tools affects what the agent can execute and how its actions are checked. \emph{SWE-agent} reports that changes to repository navigation and test-execution interfaces improve performance on software engineering tasks \citep{yang2024sweagent}. Work on tool-use alignment also studies when a model should refuse harmful instructions, distrust insecure tool responses, and avoid tool calls it can answer directly \citep{chen2024towards}. We therefore restrict worker actions through schemas, role-specific interfaces, and executable guardrails rather than relying on unconstrained model output.

The two bodies of work address complementary parts of our problem. Blockchain research supplies simulators, benchmark methods, and controlled adversarial scenarios. LLM-agent research supplies mechanisms for tool use, role separation, and feedback-guided iteration. Spec2Twin-Chain combines these elements for the narrower task of constructing and calibrating blockchain digital twins from an explicit specification and behavioral evidence.

\section{Methodology}
\label{sec:methodology}
Spec2Twin-Chain operates as a \emph{builder} for blockchain digital twins rather than a single, fixed twin instance. Given a target-system specification and behavioral evidence, the method constructs a validated digital-twin configuration by solving a constrained optimization problem. In Section~\ref{sec:bilevel-formulation}, we formulate digital-twin construction as a hierarchical optimization problem over symbolic architectures and architecture-conditioned tuning variables. In Section~\ref{sec:bilevel-solver}, we describe a bi-level solution strategy,
where an LLM proposes and refines discrete architectural decisions, while a simulation-based optimizer calibrates the associated tuning vector.

\subsection{Optimization Formulation for Digital Twin Construction}
\label{sec:bilevel-formulation}

Given a target-system specification and behavioral evidence, such as KPI summaries or time-series traces, Spec2Twin-Chain generates a digital-twin configuration that is both structurally valid and behaviorally aligned with the target system. To bridge informal requirements and executable simulation models, we rely on a \textbf{declarative structural specification}. This specification describes the admissible design space through component types, interaction patterns, abstraction levels, parameter domains, and safety constraints. 
Therefore, it guides the search toward structurally coherent digital-twin candidates and filters malformed configurations before expensive simulation.

Methodologically, we formulate twin construction as a constrained optimization problem over this structured space. The decision variables are partitioned into the twin's \emph{architecture} and its architecture-conditioned \emph{tuning variables}. The architecture captures discrete design choices, while the tuning vector captures operational behavior conditional on those choices. Specifically, let $\mathcal{S}$ be the space of candidate twin configurations blue satisfying the schema. A candidate twin $c\in\mathcal{S}$ is represented by
\begin{equation}
\label{eq:twin-configuration}
    c=(d,\theta),
\end{equation}
where:
\begin{enumerate}
    \item $d\in\mathcal{D}$ denotes the \textbf{discrete architectural variables}, in which $\mathcal D$ is the admissible architecture set. These variables define the structural composition of the twin, including:
    \begin{itemize}
        \item \textit{Component topology} $(\mathcal{G})$: the directed graph of interacting modules;
        \item \textit{Interaction templates}: the logic governing information, event, or state exchange among the modules;
        \item \textit{Fidelity or abstraction modes}: the level of behavioral details that are used to instantiate a component or interaction.
    \end{itemize}
    \item $\theta\in\Theta(d)$ denotes the \textbf{architecture-conditioned tuning variables}. These variables govern the operational behavior of the instantiated twin, including timing, rates, capacities, thresholds, budgets, protocol limits, and other model-specific controls. Many entries are continuous, but some may be bounded integer or categorical choices when such choices are part of the admissible tuning surface. The notation $\Theta(d)$ emphasizes that the dimensionality, type, and feasible domain of the tuning variables depend on the selected architecture $d$. 
    
    The practical implementation does not permit an unrestricted LLM-created tuning space. The admissible architecture set $\mathcal D$ is quantified by user-supplied requests and declarative system specifications, while final admissibility is further enforced by executable schemas and compiler-owned parameters. For reference, our current implementation contains 83 numeric parameter entries, including 46 canonical entries and 3 discrete entries.
\end{enumerate}

With the decision variables defined, digital-twin construction is formulated as the minimization of a request-conditioned objective vector. 
Let $R$ be the construction request, including the target specification, behavioral evidence, and success criteria. 
Denote by $\Omega_R=\{J_1^R,\ldots,J_m^R\}$ the objective set induced by $R$ after these requirements are translated into an executable scoring contract. The optimization problem is
\begin{equation}
\label{eq:bilevel_problem}
\begin{aligned}
    \min_{\substack{d\in\mathcal{D} \\ \theta\in\Theta(d)}}\quad
    & \mathbf{J}_R(d,\theta)=\left[J_1^R(d,\theta),\ldots,J_m^R(d,\theta)\right]^{\top} \\
    \text{s.t.}\quad
    & \mathbb{I}_{\textnormal{ct}}(d,\theta)=1, \\
    & g_k(d,\theta)\le 0,\qquad k\in\mathcal{K}.
\end{aligned}
\end{equation}
For vector-valued objective $\mathbf J_R$, its minimization problem in \eqref{eq:bilevel_problem} should be considered as a multi-objective optimization in the Pareto optimal sense. (See Section 3.2.2 and 3.2.3 for a detailed implementation recipe on Pareto optimality via feasibility-aware non-dominated principle and global Pareto archive.) In \eqref{eq:bilevel_problem}, $I_{\mathrm{ct}}$ is a structural-validity indicator and $g_k$ are dynamic guardrails evaluated during simulation. The formulation is intentionally request-conditioned: different construction tasks may instantiate different objective sets, evidence sources, and guardrail collections, while retaining the same mathematical structure. We write the problem in minimization form for compactness; objectives that are naturally maximized can be represented through equivalent loss transformations or request-specific objective conventions.

Here, $\mathbf{J}_R$ is generic.
Many construction requests draw from three recurring objective families:
\begin{enumerate}
    \item \textbf{Fit discrepancy} $(J_{\text{fit}})$ measures behavioral divergence between digital-twin outputs and reference evidence. A common form is
    \begin{equation}
    \label{eq:fit-discrepancy}
        J_{\text{fit}}(c;E)=\left\lVert \mathcal{M}(c)-\mathcal{Y}(E)\right\rVert,
    \end{equation}
    where $\mathcal M(c)$ is the vector of simulated performance measures generated by configuration $c$, and $\mathcal Y(E)$ is the evidence-derived target vector under the same measurements. Thus, each \textit{simulated} KPI is compared with the corresponding \textit{evidence} KPI, while $\mathcal M$ and $\mathcal Y$ distinguish simulator-side measurements from evidence extractions.
    The norm or distance function may be scalar, vector-valued, or distributional, depending on the request.

    \item \textbf{Stability} $(J_{\text{stab}})$ penalizes configurations that exhibit unrealistic variance, pathological operating regimes, or high failure probability. Stability objectives ensure that a candidate twin does not merely match average behavior, but also remains viable under the relevant operating conditions.

    \item \textbf{Resource cost} $(J_{\text{cost}})$ captures computational, wall-clock, or modeling burden. This objective family encourages configurations that achieve the desired behavioral fidelity without unnecessary simulation complexity.
\end{enumerate}
These families provide a useful vocabulary for constructing $\Omega_R$, but they are not mandatory dimensions of every objective vector. Some construction requests may use all three; others may use a single primary fit metric together with stability and feasibility guardrails.

The feasible space in \eqref{eq:bilevel_problem} is restricted by two mechanisms:
\begin{enumerate}
    \item \textbf{Structural validity} $(\mathbb{I}_{\textnormal{ct}})$. This static indicator enforces compliance with the declarative structural specification. For a candidate configuration $c=(d,\theta)$,
    \begin{equation}
    \label{eq:structural-validity}
    \mathbb{I}_{\textnormal{ct}}(d,\theta)=
    \begin{cases}
    1, & \text{if } d\in\mathcal{D}_{\textnormal{valid}} \text{ and } \theta\in\Theta(d),\\
    0, & \text{otherwise}.
    \end{cases}
    \end{equation}
    Here, $\mathcal{D}_{\textnormal{valid}}$ denotes schema-admissible component and interaction structures, while $\Theta(d)$ denotes the architecture-conditioned tuning domain. This check is performed before simulation and rejects malformed designs, such as incompatible modules or parameters outside admissible bounds.

    \item \textbf{Dynamic guardrails} $(g_k)$. These constraints represent runtime feasibility requirements monitored during simulation. The index set $\mathcal{K}$ may include safety, stability, conservation, budget, or resource-exhaustion constraints. A violation occurs when $g_k(d,\theta)>0$ for any $k\in\mathcal{K}$. Such candidates are either rejected, or assigned dominated objective values, consistent with standard constraint-handling principles in simulation optimization.
\end{enumerate}

\subsection{Optimization Strategy: Bi-level Decomposition}
\label{sec:bilevel-solver}
The problem in \eqref{eq:bilevel_problem} is a simulation-based mixed-variable optimization problem over symbolic structures and architecture-conditioned tuning domains. Its structure differs from classical smooth nonlinear programs in three ways:
\begin{itemize}
    \item \textbf{High-dimensional symbolic space.} The architectural variable $d$ is a symbolic configuration defined by component graphs, interaction templates, and abstraction choices. The resulting design space $\mathcal{D}$ is combinatorial and non-Euclidean, making standard relaxations or gradient-based updates unsuitable.

    \item \textbf{Stochastic and expensive black-box evaluation.} The objective vector $\mathbf{J}_R$ is evaluated through simulation. The response surface is generally non-differentiable, non-convex, noisy, and costly to sample.

    \item \textbf{Hierarchical conditionality.} The feasible tuning domain $\Theta(d)$ is conditioned on the architecture. Changing $d$ can change the dimension, interpretation, variable type, and bounds of $\theta$, so the joint space cannot be searched as a single fixed Euclidean box.
\end{itemize}

These properties motivate a \textbf{bi-level solver strategy}, which separates structural search over $d$ from parameter refinement over $\theta$. 
The original problem \eqref{eq:bilevel_problem} can be reformulated as:
\begin{equation}
\label{eq:bilevel_formulation}
\begin{aligned}
\text{(Upper Level)}\quad
& \min_{d\in\mathcal{D}}\quad
  \mathcal{P}\!\left(\mathbf{J}_R(d,\theta^*(d))\right) \\
\text{s.t.}\quad
& \mathbb{I}_{\textnormal{ct}}(d)=1, \\[5pt]
\text{(Lower Level)}\quad
& \theta^*(d)\in\arg\min_{\theta\in\Theta(d)}
  \mathbf{J}_R(d,\theta) \\
\text{s.t.}\quad
& g_k(d,\theta)\le 0,
  \qquad k\in\mathcal{K}.
\end{aligned}
\end{equation}
Here, $\theta^*(d)$ denotes the optimized tuning outcome for a fixed architecture $d$; in the multi-objective setting, this may correspond to a non-dominated set rather than a single vector. The mapping $\mathcal{P}(\cdot)$ denotes the Pareto frontier, or a request-specific summary of that frontier. Thus, the upper level compares admissible architectures by the quality of the optimized trade-off surface they induce, while the lower level searches for feasible and non-dominated parameterizations within each architecture-conditioned domain. 
However, \eqref{eq:bilevel_formulation} does not require exhaustive optimization over every $d\in\mathcal D$: the lower-level problem is solved only for the finite set of architectures proposed by the upper level and \textit{accepted by structural validation}. Computational cost is therefore proportional to the number of \textit{accepted} proposals rather than the very large $|\mathcal D|$.” For example, the reported lower-level budgets are large yet manageable: 9,920 evaluations for Part 2.1a, 3,360 for Part 2.1b, 1,344 for Part 2.2a, and 288 for Part 2.2b. (See Table 1 for task details)
With a slight abuse of notation, $\mathbb{I}_{\textnormal{ct}}(d)=1$ denotes the architecture-level checks, which do not depend on $\theta$; the $\theta$-level domain condition is enforced within the lower level.

The resulting nested solver architecture bridges semantic reasoning with simulation-based parameter search. The upper-level problem is handled by LLM-based agents, which act as heuristic architects for the symbolic design variable $d$. The lower-level problem is handled by a simulation-based optimizer that calibrates $\theta$ for each admissible architecture. This division of labor follows two complementary lines of evidence: bi-level optimization provides a principled separation between design selection and parameter calibration \citep{colson2007overview}, while tool-using scientific-agent systems show that LLMs are more reliable when they plan, retrieve, and orchestrate external tools rather than replace domain simulators or optimizers \citep{yao2023react,boiko2023coscientist,bran2024chemcrow}. The following subsections describe the two levels and the feedback mechanism that links them.

\subsubsection{Upper-Level Optimization: LLM-Driven Architectural Search}
\label{sec:upper-level}
The upper-level problem selects a discrete configuration $d\in\mathcal{D}$ that yields a high-quality optimized frontier $\mathcal{P}(\mathbf{J}_R(d,\theta^*(d)))$. This problem is difficult because the mapping from symbolic architecture to optimized performance is non-smooth. Moreover, the validity constraint $\mathbb{I}_{\textnormal{ct}}(d)=1$ can involve semantic and structural rules that are awkward to encode as ordinary integer constraints, but natural to express in a schema with language-level requirements.

LLMs are used as structured proposal mechanisms for this symbolic search. Their role is not to certify optimality. Rather, they provide informed candidate architectures, revise proposals under validation feedback, and use prior evaluated configurations to guide subsequent proposals. Each candidate remains subject to structural validation before simulation. Thus, the LLM functions as a proposal prior and interface layer, not as a substitute for formal validation or numerical optimization.

In this sense, (6) can be viewed as an instance of simulation-driven black-box optimization: candidate structures are generated without access to gradients, evaluated through an external compiler and simulation runtime, and subsequent proposals condition on previously evaluated outcomes \citep{shahriari2016taking,snoek2012practical}. Recent work that recasts black-box optimization as generative or conditional sampling includes inverse generative models \citep{kumar2020model} and diffusion-based optimization \citep{krishnamoorthy2023diffusion, li2024diffusion}. Spec2Twin-Chain is also closely related to LLM-based optimization methods such as OPRO, in which an LLM generates new candidate solutions conditioned on earlier solutions and their objective values \citep{yang2024large}. 

But unlike these previous works, our Spec2Twin-Chain maintains a clear bi-level strategy that optimizes a separate lower-level problem from the upper-level loop. Here, we first formalize the upper-level loop as follows:

\begin{enumerate}
    \item \textbf{Context initialization.} Construct a prompt context containing the construction request $R$, the target-system specification $\mathfrak{S}$, behavioral evidence $E$, the declarative schema, and the current objective contract $\mathbf{J}_R$. Similar to Retrieval-Augmented Generation (RAG), external task-specific information in our loop is supplied to the LLM at inference time, but here we perform deterministic and schema-aware retrieval rather than an unconstrained semantic document search.

    \item \textbf{Candidate generation.} At iteration $t$, the LLM proposes a set of $N$ candidate architectures
    \begin{equation}
    \label{eq:upper-level-proposals}
        \{d_t^{(1)},\ldots,d_t^{(N)}\}\sim
        P_{\textnormal{LLM}}(d\mid \mathfrak{S},E,\mathbf{J}_R,H_{t-1}),
    \end{equation}
    where $H_{t-1}$ is the feedback context constructed from previously evaluated candidates.

    \item \textbf{Schema validation.} Each candidate $d_t^{(i)}$ is checked against $\mathbb{I}_{\textnormal{ct}}$. Invalid candidates are discarded or returned to the LLM with validator diagnostics for correction. In our Part 2.1 and 2.2 experiments, all 86 upper-level proposals were processed without invoking deterministic fallbacks. In addition, Part 2.3 contains 384 schema-validated worker decisions with no fallback response. But compile-repair turns are internal to candidates, this zero-fallback statistic should not be interpreted as a zero initial schema-rejection rate.

    \item \textbf{Performance feedback.} Valid candidates are passed to the lower-level solver, which computes their optimized frontier or frontier summary. These outcomes are appended to the search history.

    \item \textbf{In-context refinement.} The history $H_t$ is formatted as a learning trajectory for the next iteration, allowing the LLM to condition future proposals on observed trade-offs among fit, stability, cost, and feasibility.
\end{enumerate}

\subsubsection{Lower-Level Optimization: Simulation-Based Parameter Refinement}
\label{sec:lower-level}
Conditioned on a valid architecture $d$, the lower-level problem becomes a simulation-based optimization problem over $\Theta(d)$. The framework does not require a unique lower-level optimizer: derivative-free methods such as evolutionary algorithms, decomposition-based multi-objective search, surrogate-assisted Bayesian optimization, or scalarized optimizers may be appropriate depending on the objective structure and evaluation budget \citep{zhang2007moead,snoek2012practical,forrester2007multi,hansen2016cma}. In this study, we instantiate the lower level with the Non-dominated Sorting Genetic Algorithm II (NSGA-II) because it maintains a population approximation to the Pareto frontier, preserves diversity through crowding distance, does not require gradients, and naturally accommodates noisy and constrained simulation outputs \citep{deb2002fast}. This choice is therefore a practical default for the multi-objective, simulation-based setting, not a restriction of the Spec2Twin-Chain formulation. Appendix~\ref{app:method-details} provides the methodological complement, including the feasibility-aware solver loop in Appendix~\ref{app:nsga2-details}, and the executable interpretation of the formulation in Appendix~\ref{app:method-executable-scope}.

Let $\mathcal{Q}_g$ denote the population of tuning vectors at generation $g$. For each $\theta_i\in\mathcal{Q}_g$, the simulator evaluates the configuration $(d,\theta_i)$ and returns the objective vector $\mathbf{J}_R(d,\theta_i)$ together with guardrail violations. Ranking follows a feasibility-aware dominance principle: feasible solutions dominate infeasible ones; infeasible solutions are ordered by their degree of constraint violation; and among feasible solutions, Pareto dominance and crowding distance guide selection. Under a finite evaluation budget, the resulting procedure returns an empirical approximation to the architecture-conditioned Pareto frontier, namely the non-dominated feasible set \citep{deb2002fast} among the evaluated points in $\Theta(d)$. Algorithm~\ref{alg:nsga2} in Appendix~\ref{app:nsga2-details} states the full loop.

\subsubsection{Global Archive and Feedback Construction}
\label{sec:archive-feedback}
The bi-level framework does not treat lower-level searches as isolated runs. Instead, it maintains a global Pareto archive $\mathcal{A}_{\textnormal{global}}$ across upper-level iterations. After computing the architecture-conditioned empirical Pareto frontier $\mathcal{P}_d^*$ for a candidate, the archive is updated by non-dominated merge:
\begin{equation}
\label{eq:global-archive-update}
\begin{aligned}
\mathcal{U}_d &= \mathcal{A}_{\textnormal{global}}\cup\mathcal{P}_d^*,\\
\mathcal{A}_{\textnormal{global}} &\leftarrow
\{c\in\mathcal{U}_d:\nexists c'\in\mathcal{U}_d,\; c'\prec c\}.
\end{aligned}
\end{equation}
where $\prec$ denotes Pareto dominance under the request-conditioned objectives and guardrail treatment.

The archive also serves as semantic memory for the upper-level architect. To construct the feedback context $H_t$, a representative subset $\mathcal{E}_t\subset\mathcal{A}_{\textnormal{global}}$ is selected. When scalar ranking is required, we use a request-conditioned utility
\begin{equation}
\label{eq:request-conditioned-utility}
U_R(d,\theta)=w^\top\widetilde{\mathbf J}_R(d,\theta),
\end{equation}
where $\widetilde{\mathbf J}_R$ is obtained by first orienting every component-wise objective to a maximization problem: we keep $z_j=J_j^R$ for maximization tasks and negate $z_j=-J_j^R$ for minimization tasks. Then we apply within-archive min–max normalization: $\widetilde J_j^R=(z_j-\min z_j)/(\max z_j-\min z_j)$.
The selected exemplars are summarized into a compact feedback context that connects design choices to realized simulation performance. This feedback construction implements in-context learning: the model receives evaluated examples from the prior search trajectory and adapts its next proposal distribution without weight updates, consistent with few-shot prompting and broader surveys of in-context learning \citep{brown2020language,dong2023survey}. We include both successful and diagnostic information (objectives, guardrail violations, and structural edits) because demonstration selection and formatting can materially affect in-context learning behavior \citep{min2022rethinking}. The archive is therefore a validated record of the search trajectory, not an unconstrained conversational log. In this way, the LLM's subsequent proposals are grounded in evaluated design trajectories. Algorithm~\ref{alg:bilevel} summarizes the complete loop.

\begin{algorithm}[ht]
\caption{Spec2Twin-Chain Bi-Level Optimization Loop}
\label{alg:bilevel}
\begin{algorithmic}[1]
\REQUIRE Request $R$ with target specification $\mathfrak{S}$ and behavioral evidence $E$; upper-level iterations $T$; candidate budget $N$
\STATE $\mathcal{A}_{\textnormal{global}}\leftarrow\emptyset$, $H_0\leftarrow\emptyset$
\FOR{$t=1$ \TO $T$}
    \STATE Infer or update objective contract $\mathbf{J}_R$ and guardrails from $R$ and $H_{t-1}$
    \STATE Generate candidate architectures $\mathcal{D}_t=\{d_t^{(1)},\ldots,d_t^{(N)}\}$ using the LLM architect
    \STATE Retain candidates satisfying $\mathbb{I}_{\textnormal{ct}}$
    \FORALL{valid $d\in\mathcal{D}_t$}
        \STATE $\mathcal{P}_d^*\leftarrow\texttt{SimOpt}(d,\mathbf{J}_R,\Theta(d),\{g_k\}_{k\in\mathcal{K}})$
        \STATE $\mathcal{A}_{\textnormal{global}}\leftarrow\texttt{NonDominatedMerge}(\mathcal{A}_{\textnormal{global}},\mathcal{P}_d^*)$
    \ENDFOR
    \STATE Select representative exemplars $\mathcal{E}_t\subset\mathcal{A}_{\textnormal{global}}$
    \STATE $H_t\leftarrow\texttt{FormatFeedback}(\mathcal{E}_t)$
\ENDFOR
\RETURN $\mathcal{A}_{\textnormal{global}}$
\end{algorithmic}
\end{algorithm}

\subsubsection{Post-Construction Optimization and Reuse}
Once a calibrated twin is available, the same hierarchical search machinery can also be reused in a distinct \textit{post-construction} stage, where the digital twin is further optimized for a downstream decision problem. This is analogous to an adaptation stage at the workflow level but different from an LLM post-training in that the model weights are not updated.
The same bi-level structure can also be used after the initial twin has been constructed. In downstream instances, the optimizer conditions on a task-specific external input $X_e$, which may be a calibrated digital twin, a controlled simulator, historical blockchain traces, protocol specifications, benchmark scenarios, or a blockchain system exposed by observation, replay, or query. The downstream problem is then defined over a new architecture or policy structure $d_e$ and associated tuning variables $\theta_e$:
\begin{equation}
\label{eq:downstream-optimization}
\min_{d_e\in\mathcal{D}_e,\; \theta_e\in\Theta_e(d_e;X_e)}
\mathbf{J}_e(d_e,\theta_e;X_e),
\end{equation}
with experiment-specific structural-validity and guardrail constraints. Under this view, a calibrated digital twin is one possible substrate rather than a required input: if $X_e$ is a built twin, the task is DT-enabled optimization; if $X_e$ is direct blockchain evidence, benchmark traces, or an observable blockchain interface, the same hierarchical search pattern supports evidence-conditioned blockchain optimization unless the task explicitly constructs a new twin. The lower-level solver is selected for the downstream task and need not be NSGA-II. It may use another simulation-optimization method suited to the objective structure, parameter domain, and evaluation budget, provided that it optimizes $\theta_e$ conditional on $d_e$ and enforces the declared guardrails.

\section{Experiments and Evaluation}
\label{sec:experiments}

We now present the numerical experiments, beginning with background on the blockchain setting for which the digital twins are built. Ethereum stages pending transactions in node-local transaction pools; these overlapping views, propagated through transaction gossip, are commonly referred to as the public mempool. A user broadcasts a transaction, it spreads across the peer-to-peer network, and it waits among other pending transactions until a block includes it. Under proof of stake, Ethereum schedules a proposer for each 12-second slot \citep{ethereumconsensus2026}. During this wait, block construction and the fee market jointly determine whether and when a transaction is included, how it is ordered, and the effective fee it pays \citep{buterin2019eip1559,roughgarden2020eip1559}. Ordering carries value of its own: a searcher that observes a pending trade can profit by submitting a competing transaction that executes ahead of it, a practice documented at scale on decentralized exchanges \citep{daian2019flash}.

This structure generates several of the protocol design problems that occupy current Ethereum research. Pending transactions in the public mempool are visible before execution, so they can be front-run, and economically valuable order flow is therefore often routed through private channels instead. The value extractable by including, excluding, or reordering transactions is known as maximal extractable value (MEV) \citep{ethereum2026mev}. MEV has motivated an out-of-protocol form of proposer-builder separation: participating validators obtain externally constructed payloads from specialized builders, which compete by submitting bids through MEV-Boost relays. Post-Merge measurement studies found that this builder-mediated path rapidly became the dominant way blocks reach the chain \citep{wahrstatter2023time,yang2024decentralization}. Countermeasures such as encrypted mempools and inclusion lists remain under debate \citep{ethereum2026pbs}. Evaluating any of these designs on the live network can have direct financial consequences, and the surrounding strategic behavior cannot be held fixed for comparison, which is what makes controlled experimentation on a calibrated model attractive.
\begin{table}[t]
\centering
\scriptsize
\renewcommand{\arraystretch}{1.0}
\caption{Experimental instances under the Spec2Twin-Chain bi-level formulation.}
\label{tab:experiment-design-summary}
\begin{tabularx}{\textwidth}{@{}p{0.15\textwidth}XX@{}}
\toprule
\textbf{Experiment} & \textbf{Substrate and purpose} & \textbf{Bi-level optimization contract} \\
\midrule
\textbf{Part 1}\par BlockSim calibration
& Ethereum proof-of-work mempool twin calibrated to BlockSim KPI evidence; tests the core construction capability of the builder.
& \textbf{$d$:} mempool, propagation, consensus, fee-market, and block-production modules.\par
  \textbf{$\theta$:} timing, rate, gas, capacity, mining, and latency variables.\par
  \textbf{Objective:} minimize KPI fit error with stability and cost terms. \\
\addlinespace[0.3em]
\textbf{Part 2.1a}\par Congestion
& Congestion-regime scenario over the calibrated mempool twin; tests whether the twin can expose queue, latency, and breakdown behavior.
& \textbf{$d$:} load-profile, queue, backpressure, bottleneck, and telemetry structure.\par
  \textbf{$\theta$:} traffic multiplier, queue threshold, backpressure coefficient, elasticity exponent, and latency sensitivity.\par
  \textbf{Objective:} match throughput behavior while monitoring latency, queue stress, and guardrails. \\
\addlinespace[0.3em]
\textbf{Part 2.1b}\par Adversarial stress
& Joint mempool--adversary stress twin over controlled adversarial scenarios; tests aggregate stress matching and class-fingerprint recovery.
& \textbf{$d$:} attack channel, adversary type, observation surface, fingerprint logic, and detection structure.\par
  \textbf{$\theta$:} attack intensity, timing, concealment, disruption, and detection-threshold variables.\par
  \textbf{Objective:} match attack profiles while balancing disruption, detectability, stability, and cost. \\
\addlinespace[0.3em]
\textbf{Part 2.2a}\par MEV-style block construction
& Validator-side block-building policy over the mempool substrate; tests controlled reward capture under a 12-second slot budget.
& \textbf{$d$:} bundle selection, conflict resolution, ordering, pruning, timeout, and fallback policy structure.\par
  \textbf{$\theta$:} search budgets, pruning thresholds, scoring weights, conflict penalties, and timing controls.\par
  \textbf{Objective:} maximize reward subject to valid-block, conflict, and slot-time constraints. \\
\addlinespace[0.3em]
\textbf{Part 2.2b}\par Inclusion policy
& Sender-side fee and replacement policy over the mempool substrate; tests reliability--cost trade-offs under controlled inclusion scenarios.
& \textbf{$d$:} adaptive fee rule, replacement strategy, urgency tiers, and target-horizon policy.\par
  \textbf{$\theta$:} fee multiplier, fee cap, replacement bump, retry interval, target horizon, and urgency threshold.\par
  \textbf{Objective:} achieve target inclusion probability while controlling fee cost, overpayment, and invalid replacements. \\
\addlinespace[0.3em]
\textbf{Part 2.3}\par GSMP-style update policy
& Update-timing supervision over a dynamic extension of the controlled Part~2.2b inclusion-policy environment, initialized from the reported sender policy and anchored to the Part~1 calibration evidence.
& \textbf{$d$:} GSMP-style update-scope structure: review clocks, validity bands, and keep, fast-update, or full-update actions.\par
  \textbf{$\theta$:} the six Part~2.2b sender-policy variables, grouped into a four-variable fast set and the full six-variable set.\par
  \textbf{Objective:} minimize an operational inclusion, fee, and stability score on locked held-out paths, reporting update activity separately. \\
\bottomrule
\end{tabularx}
\end{table}
Our evaluation targets this setting in two parts. Part~1 examines the construction problem directly: given a blockchain-system description and controlled behavioral evidence, can the builder produce a structurally valid and quantitatively calibrated mempool twin? Part~2 then studies controlled downstream uses anchored to the calibrated setting, drawn from the questions above: fee-market behavior under congestion (Part~2.1a), adversarial mempool and network conditions (Part~2.1b), controlled MEV-style block construction under a 12-second budget (Part~2.2a), sender-side fee and replacement policy (Part~2.2b), and keeping an already-optimized policy current as conditions drift (Part~2.3). These are controlled scenario packs rather than live-chain validation: each isolates whether the bi-level workflow can synthesize the required structure, tune its variables, and enforce guardrails under a known objective contract. This organization separates calibration validity from downstream usefulness.

For Part~1, we use BlockSim as the reference simulator. BlockSim is an open-source, Python-based, discrete-event blockchain simulator introduced by \citet{alharby2020blocksim}; it models blockchain systems through network, consensus, and incentive layers and supplies controlled KPI targets and traces for calibration.\footnote{BlockSim GitHub repository: \url{https://github.com/maher243/BlockSim}.} The BlockSim Ethereum model and the evidence used in Part~1 represent pre-Merge, proof-of-work Ethereum, so the Part~1 calibration targets that setting. The optimization formulation is not intrinsically tied to proof of work, although applying it to another consensus setting would require corresponding modules, evidence, and validation. The Part~2 packs are controlled scenarios, with Part~2.2a drawing on proof-of-stake era design questions; none of them establish post-Merge Ethereum, PBS, or MEV-Boost fidelity. Nor does the choice of reference simulator restrict the methodology: other open-source blockchain simulation and emulation systems, including SimBlock for block-propagation studies \citep{aoki2019simblock} and BlockEmulator for blockchain-sharding protocol emulation \citep{huang2025blockemulator}, could serve as evidence sources when their abstractions match the target specification.

Across all experiments, the upper-level design step is implemented with Codex-based LLM workers. An \emph{LLM worker} is a model instance that authors a candidate architecture, policy, or skeleton conditioned on three experiment-specific objects: 
\begin{enumerate}[label=(\alph*)]
    \item a schema, which instantiates the structural-validity condition
    $I_{\mathrm{ct}}$, determines the admissible architecture set
    $\mathcal D$ and architecture-conditioned tuning domains
    $\{\Theta(d)\}_{d\in\mathcal D}$;

    \item an objective contract, which instantiates $\mathbf J_R$,
    its own objective directions and weights, and the guardrails $\{g_k\}$;

    \item an evidence pack, which instantiates $E$.
\end{enumerate} Each accepted upper-level candidate is counted as a \emph{Codex-authored proposal}. Proposal counts refer to upper-level candidate architectures or policies that are passed to the lower-level optimizer for evaluation. Based on our implementation, for the reported Part 2 runs, \greencode{proposal_count} $=$ \greencode{algorithm_iterations} $\times$ \greencode{candidate_budget_per_iteration}. Importantly, the count does not include raw request-processing calls or within-candidate
repair turns. 

For each reported experiment, the declarative \textit{schema} fixes the admissibility rules for $\mathcal D$ and the architecture-conditioned domains $\Theta(d)$. The upper level therefore explores different $d$ within a fixed admissible universe. Separately, the
\textit{objective contract} is instantiated before search and held fixed across upper-level iterations, so archive dominance is always evaluated under a single contract. 

Table~\ref{tab:experiment-design-summary} summarizes the six experimental instances. In each case, the upper level specifies a discrete architecture, policy, or update-scope structure $d$, and the lower level optimizes architecture-conditioned tuning variables $\theta$ under objective and guardrail contracts. Appendix~\ref{app:experiment-details} provides the experiment details and additional results: runtime and reproducibility settings are reported in Section~\ref{app:experiment-runtime}; Part~1 KPI and feedback-ablation evidence is reported in Section~\ref{app:part1-calibration-details}; Part~2.1 stress-regime diagnostics are reported in Section~\ref{app:part21-stress-details}; Part~2.2 policy-optimization details are reported in Section~\ref{app:part22-policy-details}; and Part~2.3 update-policy details are reported in Section~\ref{app:part23-update-details}.

\subsection{Part 1: BlockSim-Grounded Digital-Twin Calibration}
\label{sec:exp-part1-calibration}

The first experiment evaluates the core construction capability of Spec2Twin-Chain. The target is an Ethereum proof-of-work mempool setting, and the evidence pack contains BlockSim configuration metadata, execution traces, and ten KPI targets including block production, transaction throughput, gas usage, uncle generation\footnote{An uncle (ommer) in pre-Merge proof-of-work Ethereum is a valid competing block that does not enter the canonical chain but may be referenced by a later canonical block and receive a partial reward.}, and block-time distribution. The reported evidence horizon contains 33 blocks and 4,534 transactions, corresponding to an average of 137.394 transactions per block. The ten KPIs in Figure 2 are aggregate statistics extracted from this finite horizon rather than a single block-count target. The upper-level worker constructs an EVM-PoW architecture with mempool, consensus, mining, network, execution, fee-market, state, and telemetry modules. The lower level tunes block timing, mining, gas, transaction-arrival, mempool-capacity, latency, and execution parameters within evidence-guided bounds.

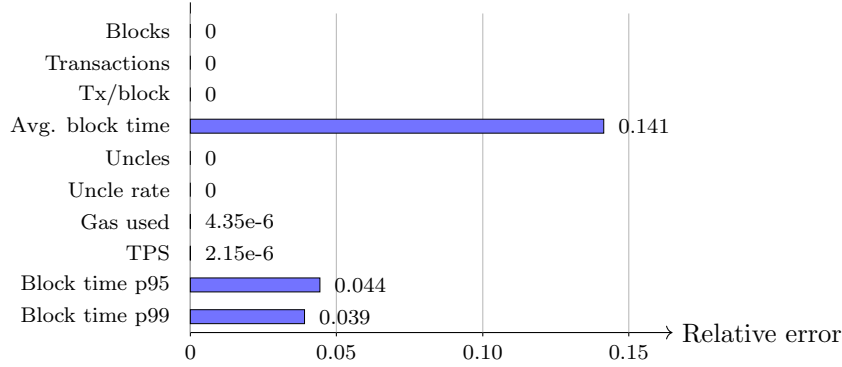
\begin{figure}[H]
\centering
\begin{tikzpicture}[x=0.58cm,y=0.42cm]
\small
\draw[->] (0,0) -- (11,0) node[right] {Relative error};
\draw (0,0) -- (0,10.4);
\foreach \x/\lab in {0/0,3.33/0.05,6.67/0.10,10/0.15} {
  \draw[gray!55] (\x,0) -- (\x,10.05);
  \draw (\x,0) -- (\x,-0.12) node[below] {\scriptsize \lab};
}
\foreach \y/\name/\val/\txt in {
  9.5/{Blocks}/0/{0},
  8.5/{Transactions}/0/{0},
  7.5/{Tx/block}/0/{0},
  6.5/{Avg. block time}/9.429/{0.141},
  5.5/{Uncles}/0/{0},
  4.5/{Uncle rate}/0/{0},
  3.5/{Gas used}/0.00029/{4.35e-6},
  2.5/{TPS}/0.00014/{2.15e-6},
  1.5/{Block time p95}/2.955/{0.044},
  0.5/{Block time p99}/2.604/{0.039}
} {
  \node[anchor=east] at (-0.25,\y) {\scriptsize \name};
  \draw[fill=blue!55,draw=black,line width=0.2pt] (0,\y-0.22) rectangle (\val,\y+0.22);
  \node[anchor=west] at (\val+0.12,\y) {\scriptsize \txt};
}
\end{tikzpicture}
\caption{KPI-level calibration error for the full-evidence BlockSim calibration run. The generated digital twin matches block count, transaction count, average transactions per block, uncle count, uncle rate, gas usage, and throughput with negligible error. The remaining aggregate fit error is concentrated in block-time distribution metrics, especially average block time.}
\label{fig:part1-kpi-fit-profile}
\end{figure}

Figure~\ref{fig:part1-kpi-fit-profile} shows the KPI-level error profile. The generated twin matches the count and throughput metrics closely: block count, transaction count, transactions per block, uncle count, uncle rate, gas usage, and TPS have zero or negligible relative error. The remaining error is concentrated in timing metrics, especially average block time. This pattern is consistent with the modeling role of the twin: the builder recovers aggregate throughput, gas, and uncle behavior while approximating the stochastic timing distribution through a lower-dimensional parameterization.

The aggregate fit error is $0.02248$, below the target threshold of $0.05$. The hard runtime guardrail is also satisfied because the calibrated average block time is 13.0683 seconds, below the 20-second limit. Therefore, the full-evidence calibration study supports the first claim: the LLM-guided upper level can produce a schema-valid architecture and scoring contract, and the lower-level optimizer can calibrate the associated tuning variables to controlled simulator evidence.

We also test whether feedback helps when the initial request is less informative. Without evidence-guided bounds, the first candidate has a fit error of $0.30079$, but the loop recovers to $0.02248$ after diagnostic feedback. Under progressive feedback, the error remains high in the first two iterations, improves after partial diagnostics, and falls below the threshold once full diagnostics are available. The detailed KPI table and ablation traces in Appendix~\ref{app:part1-calibration-details}--especially Table~\ref{tab:part1-main-kpi}, Figure~\ref{fig:part1-feedback-recovery}, and Table~\ref{tab:part1-ablation}--show that the feedback archive is not merely a log of previous evaluations; it supplies actionable signals for revising later upper-level proposals and lower-level search contracts.

Overall, Part~1 shows that Spec2Twin-Chain can construct a calibrated EVM-PoW mempool digital twin against controlled BlockSim evidence. The claim is intentionally scoped: the experiment does not imply post-Merge Ethereum fidelity, PBS or MEV-Boost fidelity, or byte-for-byte equivalence with BlockSim.

\subsection{Part 2.1: Stress-Regime Analysis with the Calibrated Mempool Twin}
\label{sec:exp-part21-stress}

Part~2.1 asks whether the calibrated mempool twin remains useful when the objective shifts from matching reference evidence to characterizing stress behavior. We consider two stress-regime scenarios. Part~2.1a uses the twin as a congestion and scaling what-if engine, while Part~2.1b uses it as an adversarial-stress substrate. In both studies, the upper level proposes a stress-specific architecture and observation surface, and the lower level tunes architecture-conditioned parameters under scenario-specific objectives and guardrails. Table~\ref{tab:part21-stress-summary} reports the compact result summary; the corresponding run-level objectives, scenario thresholds, and adversarial diagnostics are reported in Appendix~\ref{app:part21-stress-details}, Tables~\ref{tab:part21a-congestion-summary}--\ref{tab:part21b-adversarial-summary} and Figure~\ref{fig:part21b-adversarial-profile}.

\begin{table}[H]
\centering
\small
\setlength{\tabcolsep}{4pt}
\caption{Part 2.1 stress-regime results over the calibrated mempool twin.}
\label{tab:part21-stress-summary}
\begin{tabular}{p{0.16\textwidth}rrp{0.28\textwidth}p{0.24\textwidth}}
\hline
\textbf{Study} & \textbf{Proposals} & \textbf{Fallbacks} & \textbf{Primary result} & \textbf{Interpretation} \\
\hline
Part~2.1a: Congestion
& 32 & 0
& Zero throughput-fit error; conservative breakdown threshold of 76 TPS; no divergence or eviction.
& The congestion twin exposes queue and latency stress while satisfying guardrails. \\

Part~2.1b: Adversarial stress
& 6 & 0
& Zero profile-fit error; no guardrail violations; strict fingerprint match rate of 0.4.
& The adversarial twin matches aggregate stress profiles but exposes incomplete class-specific fingerprints. \\
\hline
\end{tabular}
\end{table}

\subsubsection{Part 2.1a: Congestion-Regime What-if Analysis}
\label{sec:exp-part21a-congestion}


The congestion study contains three deterministic, schema-validated stress fixtures: a peak-load TPS spike, a gas-spike trace, and a throughput-collapse trace. At each tick, a fixture specifies offered demand (\greencode{demand_tps}), block capacity (\greencode{block_capacity_tps}), baseline latency (\greencode{base_latency_ms}), and a gas multiplier (\greencode{gas_multiplier}), together with a reference throughput and guardrails. The evaluator computes the adjusted capacity as ($C_t/\max(0.25,g_t)$), propagates queue, and defines throughput fit as RMS error against the reference curve. A scenario’s breakdown threshold is the first raw offered-demand point once queue depth reaches the optimized queue threshold or the backpressure signal turns positive.

The Codex-authored congestion twin represents a queue-mediated service-compression mechanism: elevated arrivals build backlog, backlog increases fee competition and latency, and the combined pressure reduces effective inclusion capacity. Mixed-variable NSGA-II then tunes five parameters: traffic-rate multiplier, queue threshold, backpressure coefficient, elasticity exponent, and latency sensitivity.

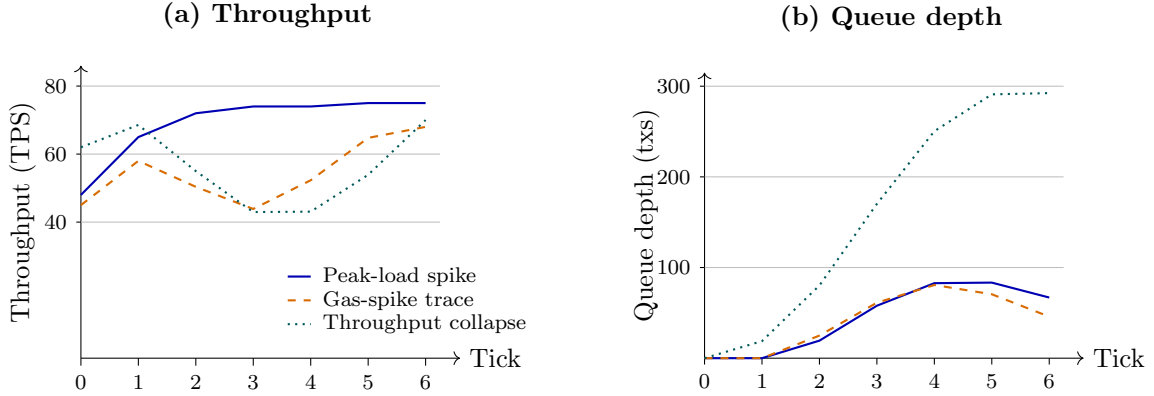
\begin{figure}[ht]
\centering
\captionsetup{width=1\textwidth}
\begin{tikzpicture}
\small
\begin{scope}[x=0.76cm,y=0.045cm]
\node[anchor=south] at (3.25,94) {\textbf{(a) Throughput}};
\draw[->] (0,0) -- (6.6,0) node[right] {Tick};
\draw[->] (0,0) -- (0,86);
\node[rotate=90,anchor=south] at (-0.62,43) {Throughput (TPS)};
\foreach \x in {0,1,2,3,4,5,6} {\draw (\x,0) -- (\x,-2.2) node[below] {\scriptsize \x};}
\foreach \y/\lab in {40/40,60/60,80/80} {\draw[gray!50] (0,\y) -- (6.25,\y); \draw (0,\y) -- (-0.08,\y) node[left] {\scriptsize \lab};}
\draw[blue!70!black,thick] (0,48.0) -- (1,65.0) -- (2,72.0) -- (3,74.0) -- (4,74.0) -- (5,75.0) -- (6,75.0);
\draw[orange!85!black,thick,dashed] (0,45.0) -- (1,58.0) -- (2,50.37037) -- (3,43.870968) -- (4,52.307692) -- (5,64.761905) -- (6,68.0);
\draw[teal!70!black,thick,dotted] (0,62.0) -- (1,68.571429) -- (2,55.0) -- (3,42.962963) -- (4,43.076923) -- (5,53.913043) -- (6,70.0);
\draw[blue!70!black,thick] (3.6,24) -- (4.0,24); \node[anchor=west] at (4.05,24) {\scriptsize Peak-load spike};
\draw[orange!85!black,thick,dashed] (3.6,17) -- (4.0,17); \node[anchor=west] at (4.05,17) {\scriptsize Gas-spike trace};
\draw[teal!70!black,thick,dotted] (3.6,10) -- (4.0,10); \node[anchor=west] at (4.05,10) {\scriptsize Throughput collapse};
\end{scope}
\begin{scope}[xshift=8.25cm,x=0.76cm,y=0.012cm]
\node[anchor=south] at (3.25,350) {\textbf{(b) Queue depth}};
\draw[->] (0,0) -- (6.6,0) node[right] {Tick};
\draw[->] (0,0) -- (0,315);
\node[rotate=90,anchor=south] at (-0.62,157.5) {Queue depth (txs)};
\foreach \x in {0,1,2,3,4,5,6} {\draw (\x,0) -- (\x,-7) node[below] {\scriptsize \x};}
\foreach \y/\lab in {100/100,200/200,300/300} {\draw[gray!50] (0,\y) -- (6.25,\y); \draw (0,\y) -- (-0.08,\y) node[left] {\scriptsize \lab};}
\draw[blue!70!black,thick] (0,0.0) -- (1,0.0) -- (2,19.34367) -- (3,57.868333) -- (4,82.744139) -- (5,83.358506) -- (6,66.895492);
\draw[orange!85!black,thick,dashed] (0,0.0) -- (1,0.0) -- (2,24.934828) -- (3,61.219274) -- (4,80.723056) -- (5,70.523707) -- (6,45.793524);
\draw[teal!70!black,thick,dotted] (0,0.0) -- (1,18.818714) -- (2,80.230925) -- (3,170.187755) -- (4,250.346962) -- (5,290.998789) -- (6,292.409456);
\end{scope}
\end{tikzpicture}
\caption{Part 2.1a congestion trajectories for the congestion what-if twin. Each trace includes an initial pre-stress operating point with zero queue and zero backpressure, providing an internal normal baseline before congestion develops. The left panel shows throughput and the right panel shows queue depth for the three controlled stress scenarios. The gas-spike trace yields the conservative breakdown threshold of 76 TPS, while the throughput-collapse trace produces the largest queue buildup and peak latency.}
\label{fig:part21a-breakdown-curves}
\end{figure}

Figure~\ref{fig:part21a-breakdown-curves} visualizes the resulting stress trajectories. The gas-spike trace gives the most conservative breakdown threshold, 76 TPS, because throughput drops while queue and latency increase. The throughput-collapse scenario produces the largest queue, 292.4095 transactions, and the peak latency, 258.8642 ms. Thus, the result is not only an aggregate objective match; it also provides an interpretable congestion profile under controlled stress conditions.

The Part~2.1a evidence supports a downstream-use claim: after calibration, Spec2Twin-Chain can reuse the mempool twin as a what-if substrate for congestion-regime analysis. The reported TPS threshold is a scenario-specific diagnostic rather than a production-chain capacity guarantee.

\subsubsection{Part 2.1b: Adversarial Stress Analysis}
\label{sec:exp-part21b-adversarial}

Part~2.1b constructs a joint mempool--adversary stress twin over five controlled attack classes: mempool spam, frontrunning injection, eclipse-style network impairment, selfish-mining timing, and covert transaction replacement. The upper-level variable defines attack channels, targets, observation surfaces, fingerprint logic, detection gates, and stability guardrails. The lower level tunes adversary budget, timing, concealment, disruption, and detection parameters.

The run reaches zero aggregate profile-fit error and no guardrail violations. However, only two of the five strict fingerprints match the scenario oracle: mempool spam and frontrunning injection. Eclipse, selfish-timing, and covert-replacement cases remain too covert or too slowly rising under the scenario criteria. This is a useful limitation rather than a hidden failure, because it separates profile-level matching from class-signature recovery. Additional high-budget probes did not improve the fingerprint rate or stability term and increased cost in one probe. Therefore, the evidence does not indicate that the remaining mismatch is only a search-budget shortage.

Overall, Part~2.1 shows that the calibrated twin can be transformed into stress-analysis twins that remain executable and guardrail-satisfying. The claims remain limited to controlled scenario analysis rather than live-chain adversarial validation.

\subsection{Part 2.2: Policy Optimization with the Calibrated Mempool Twin}
\label{sec:exp-part22-policy}

Part 2.2 is conceptually different from a digital-twin construction task. Here the calibrated mempool digital twin is fixed as a decision substrate, while the object to be optimized is a downstream policy. We therefore interpret Part 2.2 as an optimization task facilitated by one digital twin rather than as a construction of another. Admittedly, these downstream problems could be solved by other black-box optimizers, but we retain our bi-level workflow to demonstrate that the same machinery can be modified and executed even after a twin has already been established. Part~2.2a examines validator-side MEV-style block construction under a 12-second slot-time budget, and Part~2.2b examines sender-side transaction-inclusion fee policy. The question is whether the bi-level workflow can construct a policy structure $d$ and tune the associated policy variables $\theta$ to improve a controlled utility objective while preserving feasibility and guardrails. Table~\ref{tab:part22-policy-summary} summarizes the results; Appendix~\ref{app:part22-policy-details} provides the detailed MEV-style reward and slot-time visualization in Figure~\ref{fig:part22a-mev-outcome} and the inclusion-policy baseline comparison in Table~\ref{tab:part22b-inclusion-comparison}.

\begin{table}[H]
\centering
\small
\setlength{\tabcolsep}{4pt}
\caption{Part 2.2 policy-optimization summary over the calibrated mempool twin.}
\label{tab:part22-policy-summary}
\begin{tabular}{p{0.13\textwidth}rrrrp{0.36\textwidth}}
\hline
\textbf{Study} & \textbf{Proposals} & \textbf{$J_{\text{fit}}$} & \textbf{$J_{\text{stab}}$} & \textbf{$J_{\text{cost}}$} & \textbf{Main feasible outcome} \\
\hline
Part~2.2a: MEV-style block construction
& 24 & 0.0000 & 0.0000 & 4200.0000
& Full reward capture relative to the controlled reference; included bundles \texttt{alpha} and \texttt{beta}; reward gain of 1.6575 ETH; valid block in 4.2 seconds under a 12-second budget. \\

Part~2.2b: Inclusion policy
& 24 & 0.0066 & 0.0053 & 159.8878
& Minimum inclusion probability of 0.8016; total fee of 159.3803 gwei; overpayment of 0.5075 gwei; no invalid replacements. \\
\hline
\end{tabular}
\end{table}

\subsubsection{Part 2.2a: MEV-Style Block Construction Under Slot-Time Budget}
\label{sec:exp-part22a-mev}

Part~2.2a uses a controlled block-building scenario with public and private bundle candidates, archived reward values, latency terms, risk discounts, and conflict metadata. The policy family is a conflict-aware exact-subset solver: it indexes conflicts, scores adjusted value, searches feasible bundle subsets, applies deterministic tie-breaking, and enforces timeout control. The lower level tunes search-depth, candidate-limit, risk-cap, latency-penalty, surrogate-fee, and timeout parameters.

In this experiment, $J_{\mathrm{fit}}$ has a task-specific interpretation that differs from the behavioral calibration discrepancy used in Part~1. Here,
\[
J_{\mathrm{fit}}
=
\left|\Delta R-\Delta R_{\mathrm{ref}}\right|,
\]
where $\Delta R$ denotes the reward improvement achieved by the candidate policy over the baseline and $\Delta R_{\mathrm{ref}}$ denotes the reward improvement of the controlled reference outcome. Thus, $J_{\mathrm{fit}}=0$ in Part~2.2a means that the optimized policy matches the reference reward improvement. It does not indicate zero behavioral calibration error.

The optimized policy includes \texttt{bundle\_alpha} and \texttt{bundle\_beta}, rejects conflicting alternatives, and produces a valid block in 4.2 seconds under the 12-second budget. Captured value is 4.0575 ETH, compared with 2.4000 ETH for the controlled reference, giving a reward delta of 1.6575 ETH. The result shows that once the policy family is expressed as conflict-aware exact subset search, lower-level tuning can recover the scenario optimum without sacrificing valid-block or slot-time constraints. The corresponding reward and slot-time comparison is visualized in Appendix~\ref{app:part22-policy-details}, Figure~\ref{fig:part22a-mev-outcome}.

For Part~2.2a in particular, the result is not a live MEV-Boost benchmark. A real comparison against MEV-Boost or production block builders would require historical slot-level data, mempool and private-orderflow observations, relay bids, builder payloads, realized proposer values, and a same-information evaluation protocol over many slots. It would also require high-throughput EVM/state simulation and latency-aware bundle validation under proposer-time constraints. The present controlled scenario is therefore used to evaluate DT-enabled policy construction, not to claim superiority over live MEV-Boost.

\subsubsection{Part 2.2b: Transaction-Inclusion Fee Policy Optimization}
\label{sec:exp-part22b-inclusion}

Part~2.2b evaluates sender-side fee and replacement behavior across low-congestion, high-congestion, fee-spike, replacement, and target-horizon scenarios. The upper level defines an adaptive-urgency policy with a calibrated fee oracle, urgency gate, bounded replacement manager, retry scheduler, cap enforcer, and telemetry loop. The lower level tunes the base-fee multiplier, fee cap, replacement bump, retry interval, target horizon, and urgency threshold.

\begin{figure}[ht]
\centering
\includegraphics[width=0.92\textwidth]{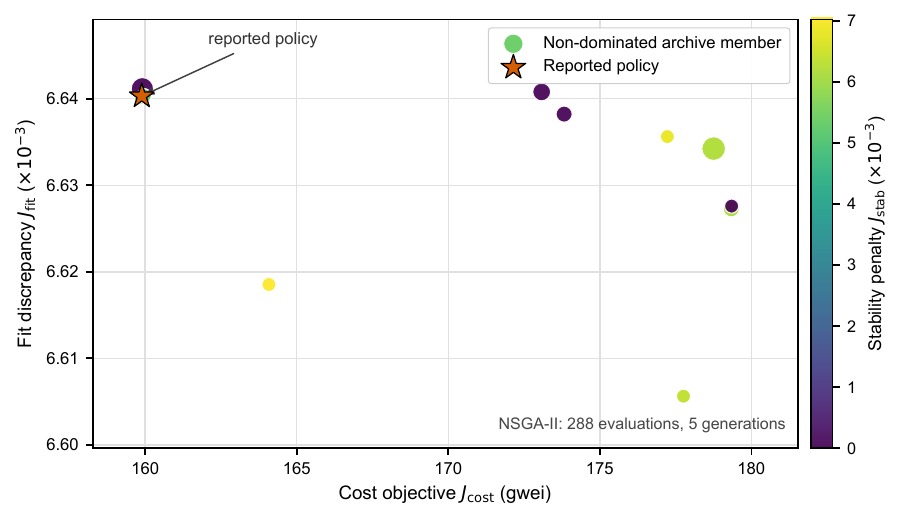}
\caption{Lower-level optimization frontier for the Part~2.2b sender-side inclusion-policy study. Each circle is a unique objective triple in the feasible run-scoped NSGA-II archive for the Codex-authored policy structure; marker size reflects repeated archive entries at the same objective values. The horizontal axis reports the cost objective $J_{\mathrm{cost}}$, the vertical axis reports fit discrepancy $J_{\mathrm{fit}}$, and color reports the stability penalty $J_{\mathrm{stab}}$. The highlighted solution is the policy reported in Table~\ref{tab:part22-policy-summary}.}
\label{fig:part22b-inner-frontier}
\end{figure}

Figure~\ref{fig:part22b-inner-frontier} exposes the lower-level optimization result behind the selected policy. Each circle is a unique objective triple from the feasible run-scoped NSGA-II archive for the fixed Codex-authored policy structure, with marker size reflecting repeated archive entries at the same objective values. The archive records 29 feasible entries, corresponding to 13 unique objective triples, after 288 lower-level evaluations and five completed generations. The reported policy is the lowest-cost feasible archive member: it clears the minimum inclusion-probability target with a worst-case inclusion probability of 0.8016, pays 159.3803 gwei, and incurs only 0.5075 gwei of overpayment. The full baseline comparison and tuned-policy summary are reported in Appendix~\ref{app:part22-policy-details}, Table~\ref{tab:part22b-inclusion-comparison}. In this controlled setting, the result shows how a calibrated digital twin can support policy comparison.

\subsection{Part 2.3: GSMP-Style Update Timing for the Optimized Inclusion Policy}
\label{sec:exp-part23-gsmp-update}

Parts~1 and~2.2 provide a calibrated mempool setting and a tuned sender-side inclusion policy. Part~2.3 asks what should happen to that policy afterwards. As transaction, block, and fee observations arrive, re-optimizing the deployed policy at every observation can chase transient noise and destabilize behavior that was already acceptable. The study therefore treats update timing itself as the decision problem. It extends the controlled Part~2.2b inclusion-policy environment along locked dynamic paths, initializes every arm from the reported sender policy, and uses the Part~1 results as calibration evidence for the broader mempool setting.

We model the update decision through a GSMP-style event-clock representation of the policy's operating environment \citep{glasserman1992gsmp}; the correspondence to the GSMP formalism is given in Appendix~\ref{app:part23-update-details}. The discrete scheme consists of transaction, block, inclusion, replacement, and review events. Its clock state records hourly and daily reviews, persistence outside the validity band, retry timing, and time since the last accepted update. At each review the supervisor also observes rolling base-fee and mempool-pressure estimates and recent inclusion, fee, and replacement outcomes. The daily review supersedes the hourly one when the two coincide. The action set is deliberately small: keep the current parameters, apply a fast update that re-optimizes the four fast policy variables, or apply a full update of all six variables, with full updates available only at daily reviews. Four update policies are compared under this contract: never update, always update, a rule-based selective baseline whose state-to-action mapping was tuned on development seeds and then frozen, and LLM-guided selective updating, in which a live worker chooses the action at each review. All four start from the same reported Part~2.2b parameter vector, observe the same event paths under common random numbers, and use the same lower-level optimizer, bounds, and per-invocation budget whenever re-optimization is requested. Every protocol choice was frozen before the locked evaluation, in which the worker made 384 review decisions, one at each review, without access to hidden regime labels or future events. Appendix~\ref{app:part23-update-details} records the full protocol.

The objective here is user-specified rather than an instance of the construction families in Section~\ref{sec:bilevel-formulation}, so we give it its own symbol. The operational score $J_{\mathrm{op}}$ combines inclusion reliability at the target horizon, fee and overpayment, and the stability of realized inclusion performance. It deliberately excludes optimizer, worker, and update costs. An always-updating policy is therefore not charged for the act of updating and can lose only through worse realized outcomes; update activity is reported separately.

\begin{table}[H]
\centering
\small
\setlength{\tabcolsep}{5pt}
\caption{Part 2.3 locked update-policy comparison over eight held-out seeds. Lower $J_{\mathrm{op}}$ is better. Updates per day counts accepted parameter changes; optimizer launches, including refits that returned no accepted change, are reported in Appendix~\ref{app:part23-update-details}.}
\label{tab:part23-update-summary}
\begin{tabular}{lrrr}
\hline
\textbf{Update policy} & \textbf{$J_{\mathrm{op}}$} & \textbf{Mean inclusion prob.} & \textbf{Updates/day} \\
\hline
Never update & 0.0974 & 0.9370 & 0.0000 \\
Rule-based selective & 0.0950 & 0.9440 & 1.4375 \\
Always update & 0.0996 & 0.9342 & 15.1875 \\
LLM-guided selective & 0.0924 & 0.9411 & 1.0625 \\
\hline
\end{tabular}
\end{table}

\begin{figure}[H]
\centering
\begin{tikzpicture}[x=0.26cm,y=0.05cm]
\small
\fill[orange!18] (2.5,0) rectangle (5.5,66);
\fill[orange!18] (8.5,0) rectangle (11.5,66);
\fill[orange!18] (14.5,0) rectangle (17.5,66);
\fill[red!10] (20.5,0) rectangle (40.5,66);
\fill[teal!10] (40.5,0) rectangle (48.5,66);
\draw[gray!50] (0,20) -- (48.5,20);
\draw[gray!50] (0,40) -- (48.5,40);
\draw[gray!50] (0,60) -- (48.5,60);
\draw[->] (0,0) -- (50.5,0) node[right] {Hour};
\draw[->] (0,0) -- (0,80);
\node[rotate=90,anchor=south] at (-1.9,33) {Rolling base fee (gwei)};
\foreach \x in {6,12,18,24,30,36,42,48} {\draw (\x,0) -- (\x,-2.4) node[below] {\scriptsize \x};}
\foreach \y in {20,40,60} {\draw (0,\y) -- (-0.5,\y) node[left] {\scriptsize \y};}
\draw[gray!60,dashed] (24,0) -- (24,66);
\draw[gray!60,dashed] (48,0) -- (48,66);
\draw[blue!70!black,thick]
(1,10.0) -- (2,9.8) -- (3,24.7) -- (4,24.6) -- (5,23.1) -- (6,10.1) -- (7,10.0) -- (8,10.0) -- (9,25.1) -- (10,23.7) -- (11,25.8) -- (12,10.1) -- (13,10.3) -- (14,9.7) -- (15,25.0) -- (16,25.2) -- (17,24.6) -- (18,9.9) -- (19,11.1) -- (20,10.6) -- (21,57.5) -- (22,26.6) -- (23,48.2) -- (24,57.2) -- (25,26.8) -- (26,52.0) -- (27,60.5) -- (28,27.2) -- (29,46.5) -- (30,58.9) -- (31,26.2) -- (32,49.3) -- (33,58.3) -- (34,28.0) -- (35,47.7) -- (36,62.9) -- (37,27.3) -- (38,49.7) -- (39,55.8) -- (40,27.5) -- (41,10.4) -- (42,9.6) -- (43,9.9) -- (44,9.9) -- (45,10.3) -- (46,10.3) -- (47,9.9) -- (48,10.1);
\draw[->,very thick,teal!60!black] (25,74) -- (25,31);
\node[anchor=south] at (25,74) {\scriptsize fast update};
\draw[->,very thick,teal!60!black] (45,26) -- (43.3,13);
\node[anchor=south] at (45,26) {\scriptsize recovery retune};
\node[anchor=south] at (10,66.5) {\scriptsize temporary shocks (kept)};
\node[anchor=south] at (33.5,66.5) {\scriptsize persistent change};
\node[anchor=south] at (45.5,66.5) {\scriptsize recovery};
\end{tikzpicture}
\caption{Part 2.3 review trace for one locked seed under LLM-guided selective updating. The line is the rolling base-fee estimate observed at each review; shaded bands are the hidden scenario periods, used only for post-run classification and never shown to the worker; dashed vertical lines mark daily reviews. The worker keeps the deployed parameters through three temporary fee shocks, accepts one targeted fast update once mempool conditions have remained outside the validated band, and accepts one retune after conditions return to the band.}
\label{fig:part23-update-timeline}
\end{figure}
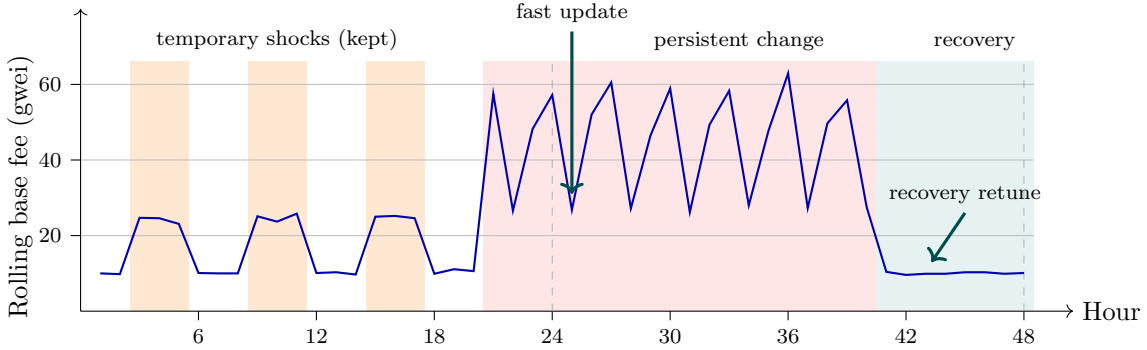

Table~\ref{tab:part23-update-summary} reports the locked comparison over eight held-out seeds. LLM-guided selective updating attains the best operational score with about one accepted update per day. The paired differences against each baseline favor the LLM policy on all eight seeds, and their 95\% confidence intervals exclude zero (Appendix~\ref{app:part23-update-details}, Table~\ref{tab:part23-paired-comparisons}). Figure~\ref{fig:part23-update-timeline} shows the mechanism on one locked seed. The worker keeps the deployed parameters through three short fee shocks, requests one targeted fast update after mempool conditions have remained outside the validated band, and applies one retune after conditions return to the band. Never updating avoids churn but concedes inclusion reliability once conditions shift persistently. Always updating performs worst despite paying no update penalty. Its repeated refits use short trailing windows, and the resulting pattern is consistent with overreaction to transient spikes.

The claim is scoped to this controlled setting: a dynamic extension of the Part~2.2b inclusion-policy evaluator, a finite update-policy set, and locked synthetic event paths. The evidence supports selective, evidence-conditioned update timing, and it shows a measurable contribution from the live worker beyond the frozen rule-based supervisor. It does not establish a universal update cadence and does not validate a live Ethereum fee policy.

Taken together, Part~2 shows how calibration evidence can support controlled stress analysis, policy optimization, and online policy-maintenance tasks. The downstream routines screen candidates against declared guardrails, expose diagnostic limitations, and produce interpretable decision outputs. These experimental claims remain scoped to controlled simulators and curated benchmark scenarios, which is appropriate before live-system validation.

\section{Conclusion and Future Work}
\label{sec:conclusion}

This study introduced \textit{Spec2Twin-Chain}, a methodology for constructing blockchain digital twins through a request-conditioned bi-level optimization framework. Rather than treating a digital twin as a fixed simulator instance, the proposed framework formulates twin construction as the joint selection of architecture-level design variables and architecture-conditioned tuning variables under structural validity and operational guardrails. The upper level uses LLM-based architectural search to translate system specifications into candidate twin structures, while the lower level uses simulation-based optimization to refine each candidate against behavioral objectives. The experiments show how this framework can calibrate a BlockSim-grounded mempool twin and then use controlled downstream environments anchored to that calibration for stress-regime analysis, policy optimization, and update-timing studies. Across these studies, the main contribution is not a single blockchain simulator, but a structured workflow that connects specification, synthesis, calibration, validation, and reuse within one optimization-oriented digital-twin building process.

Future work can extend this framework in three directions. First, empirical validation can move from controlled scenarios toward richer historical blockchain data. This is particularly important for MEV- and PBS-related studies, where recent measurement work shows that credible evaluation requires relay bids, builder payloads, proposer outcomes, slot timing, and private-orderflow information rather than finalized-chain data alone \citep{wahrstatter2023time,yang2024decentralization}. Second, the lower-level optimizer can be made adaptive to the evaluation budget and objective structure. The present experiments use NSGA-II for the constrained multi-objective construction and downstream studies in Parts~1--2.2, while Part~2.3 uses bounded random search for a smaller per-review problem. Future implementations could compare these choices with decomposition-based search, Bayesian optimization, surrogate-assisted evolutionary algorithms, and CMA-ES-style methods when simulations become more expensive or when only a limited number of evaluations is available \citep{zhang2007moead,snoek2012practical,forrester2007multi,hansen2016cma}. Third, the feedback loop can be extended from episodic archive-based refinement to online recalibration with uncertainty quantification. Digital-twin research emphasizes that trustworthy twins require not only model updating, but also explicit treatment of uncertainty across data, calibration, prediction, and optimization \citep{diamantopoulos2022dynamic,thelen2023comprehensive}. Incorporating these mechanisms would allow Spec2Twin-Chain to report when a calibrated twin remains reliable, when it has drifted outside its validated operating regime, and when new evidence should trigger reconstruction or human review.

\newpage
\bibliographystyle{apalike}
\bibliography{sample}

\newpage

\appendix

\section{Experiment Details and Additional Results}
\label{app:experiment-details}

\subsection{Experimental Runtime and Reproducibility Details}
\label{app:experiment-runtime}

This appendix records the run settings that are summarized in Section~\ref{sec:experiments}. Proposal counts refer to accepted Codex-authored upper-level candidates evaluated by the methodology runtime. Fallbacks denote deterministic candidate designs injected by the runtime when an LLM-generated proposal is unavailable or invalid. In the reported Part~2 runs, no fallback candidate is used. For Part~2.3, the proposal column instead counts bounded per-review update-scope decisions by the live worker; each is a schema-validated worker call that selects an update action, not a candidate architecture.

\begin{table}[H]
\centering
\small
\setlength{\tabcolsep}{4pt}
\caption{Runtime settings for the reported experiment instances.}
\label{tab:app-runtime-settings}
\begin{tabular}{p{0.16\textwidth}rrp{0.28\textwidth}p{0.25\textwidth}}
\hline
\textbf{Experiment} & \textbf{Proposals} & \textbf{Fallbacks} & \textbf{Lower-level setting} & \textbf{Termination / budget note} \\
\hline
Part~1 calibration
& -- & --
& Population 160; 80 optimizer generations; Stage-2 lower-level budget of 20 iterations
& Up to three upper-level construction iterations under full BlockSim evidence. \\

Part~2.1a congestion
& 32 & 0
& Mixed-variable NSGA-II; population 320; maximum 240 generations
& Plateau after 9{,}920 evaluations and 30 completed generations. \\

Part~2.1b adversarial stress
& 6 & 0
& Mixed-variable NSGA-II; population 160; maximum 100 generations
& Plateau after 3{,}360 evaluations and 20 generations. \\

Part~2.2a MEV-style block construction
& 24 & 0
& Mixed-variable NSGA-II; population 64
& Plateau after 1{,}344 evaluations and 20 generations. \\

Part~2.2b inclusion policy
& 24 & 0
& Mixed-variable NSGA-II; population 48
& Plateau after 288 evaluations and five generations. \\

Part~2.3 update policy
& 384 & 0
& Seeded bounded random search; 64 candidates per launch; trailing three-hour calibration window
& Fixed 48-review paths on eight locked seeds; feasible improvement required; no plateau rule. \\
\hline
\end{tabular}
\end{table}

\subsection{Part 1 Calibration and Feedback Details}
\label{app:part1-calibration-details}

Table~\ref{tab:part1-main-kpi} reports the full KPI comparison for the Part~1 calibration experiment. The aggregate fit error is the mean relative discrepancy over the ten comparable KPIs. The timing metrics account for most of the residual error, while count, throughput, gas, and uncle metrics are matched exactly or nearly exactly.

\begin{table}[H]
\centering
\small
\setlength{\tabcolsep}{5pt}
\caption{Part 1 full-evidence calibration run: KPI comparison between BlockSim evidence and the generated digital twin.}
\label{tab:part1-main-kpi}
\begin{tabular}{lrrr}
\hline
\textbf{KPI} & \textbf{BlockSim} & \textbf{Digital twin} & \textbf{Relative error} \\
\hline
Blocks & 33 & 33 & 0.0000 \\
Total transactions & 4534.0000 & 4534.0000 & 0.0000 \\
Transactions per block & 137.3939 & 137.3939 & 0.0000 \\
Average block time & 15.2211 & 13.0683 & 0.1414 \\
Total uncle count & 9.0000 & 9.0000 & 0.0000 \\
Uncle rate & 0.2727 & 0.2727 & 0.0000 \\
Total gas used & 199106942.0000 & 199106076.0000 & 0.000004 \\
Throughput (TPS) & 9.3086 & 9.3086 & 0.000002 \\
Block time, 95th percentile & 33.9403 & 32.4357 & 0.0443 \\
Block time, 99th percentile & 35.3770 & 36.7589 & 0.0391 \\
\hline
Aggregate fit error & \multicolumn{3}{r}{0.022484} \\
Target threshold & \multicolumn{3}{r}{0.050000} \\
\hline
\end{tabular}
\end{table}

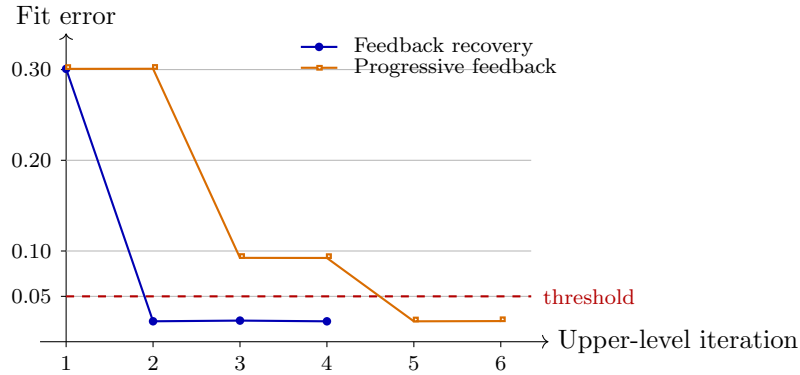
\begin{figure}[H]
\centering
\begin{tikzpicture}[x=1.15cm,y=12cm]
\small
\draw[->] (0.7,0) -- (6.55,0) node[right] {Upper-level iteration};
\draw[->] (1,0) -- (1,0.34) node[above] {Fit error};
\foreach \x in {1,2,3,4,5,6} {
  \draw (\x,0) -- (\x,-0.006) node[below] {\scriptsize \x};
}
\foreach \y/\lab in {0.05/0.05,0.10/0.10,0.20/0.20,0.30/0.30} {
  \draw[gray!55] (1,\y) -- (6.35,\y);
  \draw (1,\y) -- (0.94,\y) node[left] {\scriptsize \lab};
}
\draw[red!70!black,dashed,thick] (1,0.05) -- (6.35,0.05);
\node[anchor=west,red!70!black] at (6.38,0.05) {\scriptsize threshold};
\draw[blue!70!black,thick] (1,0.300786) -- (2,0.022520) -- (3,0.023280) -- (4,0.022484);
\foreach \x/\y in {1/0.300786,2/0.022520,3/0.023280,4/0.022484} {
  \fill[blue!70!black] (\x,\y) circle (1.6pt);
}
\draw[orange!85!black,thick] (1,0.300786) -- (2,0.300902) -- (3,0.092440) -- (4,0.092336) -- (5,0.022486) -- (6,0.022753);
\foreach \x/\y in {1/0.300786,2/0.300902,3/0.092440,4/0.092336,5/0.022486,6/0.022753} {
  \draw[orange!85!black,fill=white,thick] (\x,\y) rectangle +(0.045,0.004);
}
\draw[blue!70!black,thick] (3.7,0.325) -- (4.12,0.325); \fill[blue!70!black] (3.91,0.325) circle (1.6pt);
\node[anchor=west] at (4.2,0.325) {\scriptsize Feedback recovery};
\draw[orange!85!black,thick] (3.7,0.302) -- (4.12,0.302); \draw[orange!85!black,fill=white,thick] (3.89,0.300) rectangle (3.94,0.304);
\node[anchor=west] at (4.2,0.302) {\scriptsize Progressive feedback};
\end{tikzpicture}
\caption{Feedback recovery behavior under weakened request conditions. When evidence-guided bounds are removed, the initial candidate is poorly calibrated. Diagnostic feedback allows the bi-level loop to recover below the target fit-error threshold of 0.05. The progressive-feedback setting recovers more slowly because detailed diagnostic information is deliberately delayed.}
\label{fig:part1-feedback-recovery}
\end{figure}

The feedback studies weaken the initial request by removing evidence-guided parameter bounds. Figure~\ref{fig:part1-feedback-recovery} and Table~\ref{tab:part1-ablation} report the resulting recovery behavior. With full diagnostic feedback, the loop reaches the calibrated fit level by the fourth iteration. With progressive feedback, recovery is slower, but the final fit error still falls below the target threshold.

\begin{table}[H]
\centering
\small
\setlength{\tabcolsep}{5pt}
\caption{Part 1 feedback ablations. Both ablations begin from weaker request conditions than the full-evidence calibration run.}
\label{tab:part1-ablation}
\begin{tabular}{p{0.24\textwidth}p{0.36\textwidth}rr}
\hline
\textbf{Study} & \textbf{Condition} & \textbf{Initial fit error} & \textbf{Best fit error} \\
\hline
Full-evidence calibration run
& Full BlockSim evidence with evidence-guided parameter bounds
& -- & 0.022484 \\

Feedback-recovery ablation
& Weakened request without evidence-guided bounds; full diagnostics after first iteration
& 0.300786 & 0.022484 \\

Progressive-feedback ablation
& Weakened request without evidence-guided bounds; diagnostic feedback revealed gradually
& 0.300786 & 0.022486 \\
\hline
\end{tabular}
\end{table}

\subsection{Part 2.1 Stress-Regime Details}
\label{app:part21-stress-details}

Table~\ref{tab:part21a-congestion-summary} reports the objective and runtime metrics for the congestion-regime experiment. The optimized congestion parameters are a traffic multiplier of 1.0000, queue threshold of 1750.522 transactions, backpressure coefficient of 0.6244, elasticity exponent of 1.6862, and latency sensitivity of 0.1511.

\begin{table}[H]
\centering
\small
\setlength{\tabcolsep}{5pt}
\caption{Part 2.1a congestion-regime run. Objective values and runtime metrics are reported for the Codex-authored architecture after lower-level optimization.}
\label{tab:part21a-congestion-summary}
\begin{tabular}{lr}
\hline
\textbf{Quantity} & \textbf{Value} \\
\hline
Codex-authored upper-level proposals & 32 \\
Worker fallbacks & 0 \\
Lower-level evaluations used & 9{,}920 \\
$J_{\text{fit}}$ & 0.0000 \\
$J_{\text{stab}}$ & 1254.4253 \\
$J_{\text{cost}}$ & 84.4565 \\
Conservative breakdown threshold & 76 TPS \\
RMS throughput error & 0.0000 \\
Stress-point throughput & 42.9630 TPS \\
Stress-point latency & 258.8642 ms \\
Stress-point queue depth & 292.4095 txs \\
Eviction rate & 0.0000 \\
Diverged & No \\
Guardrails satisfied & Yes \\
\hline
\end{tabular}
\end{table}

Table~\ref{tab:part21a-scenario-thresholds} reports the scenario-level thresholds. The conservative threshold used in the main text is the minimum threshold across the three controlled stress scenarios.

\begin{table}[H]
\centering
\small
\setlength{\tabcolsep}{4pt}
\caption{Scenario-level congestion thresholds for Part 2.1a.}
\label{tab:part21a-scenario-thresholds}
\begin{tabular}{lrrrr}
\hline
\textbf{Stress scenario} & \textbf{Threshold} & \textbf{Min throughput} & \textbf{Max queue} & \textbf{Peak latency} \\
 & \textbf{(TPS)} & \textbf{(TPS)} & \textbf{(txs)} & \textbf{(ms)} \\
\hline
Peak-load TPS spike & 92.0 & 48.0000 & 83.3585 & 145.0698 \\
Gas-spike trace & 76.0 & 43.8710 & 80.7231 & 232.6230 \\
Throughput-collapse trace & 88.0 & 42.9630 & 292.4095 & 258.8642 \\
\hline
\end{tabular}
\end{table}

\begin{figure}[H]
\centering
\begin{tikzpicture}[x=1.10cm,y=3.2cm]
\small
\draw[->] (0,0) -- (6.05,0) node[right] {Diagnostic};
\draw[->] (0,0) -- (0,1.12) node[above] {Score};
\foreach \y/\lab in {0.25/0.25,0.50/0.50,0.75/0.75,1.00/1.00} {
  \draw[gray!50] (0,\y) -- (5.75,\y);
  \draw (0,\y) -- (-0.05,\y) node[left] {\scriptsize \lab};
}
\draw[fill=blue!60,draw=black,line width=0.2pt] (0.52,0) rectangle (1.08,1.0000);
\node[anchor=north,align=center] at (0.8,-0.04) {\scriptsize Profile\\[-1pt]\scriptsize fit};
\node[anchor=south] at (0.8,1.0200) {\scriptsize 1.0000};
\draw[fill=orange!80,draw=black,line width=0.2pt] (1.82,0) rectangle (2.38,0.4000);
\node[anchor=north,align=center] at (2.1,-0.04) {\scriptsize Finger-\\[-1pt]\scriptsize print};
\node[anchor=south] at (2.1,0.4200) {\scriptsize 0.4000};
\draw[fill=teal!65,draw=black,line width=0.2pt] (3.12,0) rectangle (3.68,0.3384);
\node[anchor=north,align=center] at (3.4,-0.04) {\scriptsize Detect-\\[-1pt]\scriptsize ability};
\node[anchor=south] at (3.4,0.3584) {\scriptsize 0.3384};
\draw[fill=green!55!black,draw=black,line width=0.2pt] (4.42,0) rectangle (4.98,1.0000);
\node[anchor=north,align=center] at (4.7,-0.04) {\scriptsize Guard-\\[-1pt]\scriptsize rails};
\node[anchor=south] at (4.7,1.0200) {\scriptsize 1.0000};
\end{tikzpicture}
\caption{Part 2.1b adversarial-stress diagnostics. The run obtains exact aggregate profile fit and satisfies guardrails, but only two of the five strict class fingerprints match the scenario oracle. This partial fingerprint match is treated as a diagnostic limitation rather than hidden by the aggregate objective.}
\label{fig:part21b-adversarial-profile}
\end{figure}
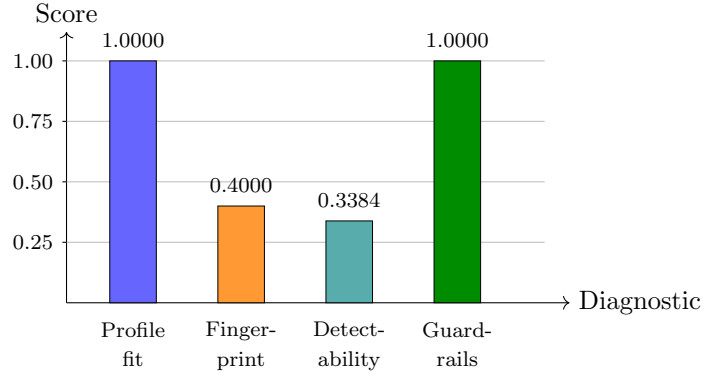

Table~\ref{tab:part21b-adversarial-summary} and Figure~\ref{fig:part21b-adversarial-profile} report the adversarial-stress diagnostics. The key distinction is between aggregate profile fit and strict class-fingerprint recovery: the aggregate profile fits exactly, but only two of five strict scenario fingerprints match. At the class level, the mempool-spam and frontrunning-injection scenarios satisfy the strict fingerprint oracle, whereas the eclipse-style, selfish-timing, and covert-replacement scenarios remain below the strict class-signature criteria.

\begin{table}[H]
\centering
\small
\setlength{\tabcolsep}{5pt}
\caption{Part 2.1b adversarial-stress run. The run covers five controlled attack classes and reports objective values after lower-level optimization.}
\label{tab:part21b-adversarial-summary}
\begin{tabular}{lr}
\hline
\textbf{Quantity} & \textbf{Value} \\
\hline
Codex-authored upper-level proposals & 6 \\
Worker fallbacks & 0 \\
Lower-level evaluations used & 3{,}360 \\
Attack classes covered & 5 \\
Attack-frontier rows emitted & 80 \\
$J_{\text{fit}}$ & 0.0000 \\
$J_{\text{stab}}$ & 0.3318 \\
$J_{\text{cost}}$ & 169.1918 \\
Profile-fit error & 0.0000 \\
Strict fingerprint match rate & 0.4000 \\
Mean detectability score & 0.3384 \\
Mean detection latency & 6.6000 blocks \\
Guardrail violations & 0 \\
Guardrails satisfied & Yes \\
\hline
\end{tabular}
\end{table}

\subsection{Part 2.2 Policy-Optimization Details}
\label{app:part22-policy-details}

\begin{figure}[H]
\centering
\begin{tikzpicture}
\small
\begin{scope}[x=1.0cm,y=0.78cm]
\node[anchor=west] at (0,5.05) {\textbf{(a) Captured reward}};
\draw[->] (0,0) -- (3.4,0);
\draw[->] (0,0) -- (0,4.65);
\node[rotate=90,anchor=south] at (-0.55,2.30) {\scriptsize ETH};
\foreach \y/\lab in {1/1,2/2,3/3,4/4} {\draw[gray!50] (0,\y) -- (3.1,\y); \draw (0,\y) -- (-0.05,\y) node[left] {\scriptsize \lab};}
\draw[fill=gray!45,draw=black,line width=0.2pt] (0.65,0) rectangle (1.25,2.4);
\draw[fill=blue!60,draw=black,line width=0.2pt] (1.9,0) rectangle (2.5,4.0575);
\node[anchor=north,align=center] at (0.95,-0.08) {\scriptsize Reference};
\node[anchor=north,align=center] at (2.2,-0.08) {\scriptsize Optimized};
\node[anchor=south] at (0.95,2.47) {\scriptsize 2.4000};
\node[anchor=south] at (2.2,4.13) {\scriptsize 4.0575};
\end{scope}

\begin{scope}[xshift=5.4cm,x=0.34cm,y=0.74cm]
\node[anchor=west] at (0,5.35) {\textbf{(b) Slot-time budget}};
\draw[->] (0,0) -- (13,0) node[right] {seconds};
\draw (0,0) -- (0,3.0);
\foreach \x/\lab in {0/0,4/4,8/8,12/12} {\draw[gray!50] (\x,0) -- (\x,2.65); \draw (\x,0) -- (\x,-0.08) node[below] {\scriptsize \lab};}
\draw[fill=blue!60,draw=black,line width=0.2pt] (0,1.9) rectangle (4.2,2.35);
\draw[fill=gray!35,draw=black,line width=0.2pt] (0,0.75) rectangle (12.0,1.2);
\node[anchor=west] at (4.35,2.12) {\scriptsize Used: 4.2 s};
\node[anchor=west] at (12.15,0.98) {\scriptsize Budget: 12 s};
\draw[red!70!black,dashed,thick] (12,0.35) -- (12,2.7);
\end{scope}
\end{tikzpicture}
\caption{Part 2.2a MEV-style block-construction outcome. The optimized conflict-aware exact-subset policy captures the full reward improvement relative to the controlled reference while satisfying the valid-block and 12-second slot-time constraints.}
\label{fig:part22a-mev-outcome}
\end{figure}
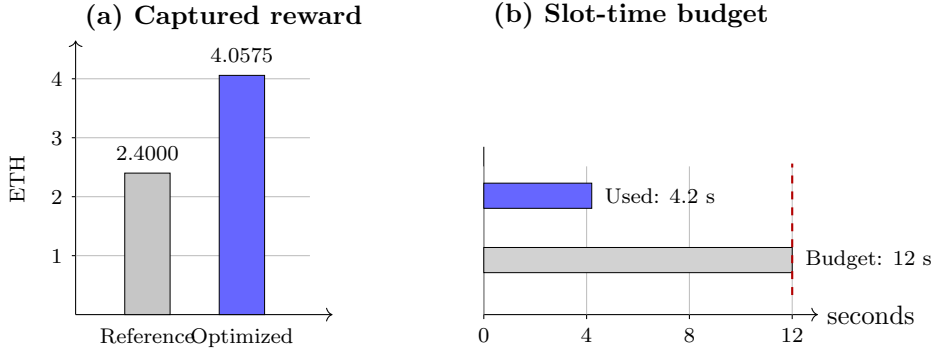

Figure~\ref{fig:part22a-mev-outcome} reports the reward and slot-time outcomes for the MEV-style block-construction scenario. The optimized conflict-aware exact-subset policy includes \texttt{bundle\_alpha} and \texttt{bundle\_beta}, excludes the conflicting alternatives, and remains inside the 12-second application budget. The tuned block-construction policy uses the largest allowed exact-subset and bundle-candidate limits in the controlled scenario, assigns zero latency penalty for this deterministic slot instance, and reserves no additional timeout slack because the constructed block remains well inside the 12-second budget.

Table~\ref{tab:part22b-inclusion-comparison} reports the inclusion-policy baselines. The optimized sender clears the target reliability threshold with substantially less fee expenditure than the conservative high-fee baseline. The optimized sender parameters are a base-fee multiplier of 1.2507, replacement bump of 18.3833\%, retry interval of 50{,}475.8 ms, target horizon of 6.5920 blocks, and urgency threshold of 0.7533. These values implement a wait-under-normal-conditions policy that escalates only when inclusion reliability approaches the target boundary.

\begin{table}[H]
\centering
\small
\setlength{\tabcolsep}{5pt}
\caption{Part 2.2b inclusion-policy comparison. The optimized policy is compared with naive low-fee and conservative high-fee baselines under the same controlled scenario set.}
\label{tab:part22b-inclusion-comparison}
\begin{tabular}{lrrrrr}
\hline
\textbf{Policy} & \textbf{Min incl.} & \textbf{Mean incl.} & \textbf{Fee} & \textbf{Overpay} & \textbf{Target sat.} \\
 & \textbf{prob.} & \textbf{prob.} & \textbf{(gwei)} & \textbf{(gwei)} & \textbf{rate} \\
\hline
Naive low-fee sender & 0.7404 & 0.8422 & 156.6336 & 0.0000 & 0.4000 \\
Conservative high-fee sender & 0.9553 & 0.9878 & 392.1200 & 109.0000 & 1.0000 \\
Codex + NSGA-II optimized sender & 0.8016 & 0.9087 & 159.3803 & 0.5075 & 1.0000 \\
\hline
\end{tabular}
\end{table}

\subsection{Part 2.3 Update-Policy Details}
\label{app:part23-update-details}

The update-timing study uses disjoint seed sets. Three development-tuning seeds were used to design the dynamic paths, observable state, worker prompt, parameter grouping, optimizer settings, and rule-based baseline. Three additional development-validation seeds were used to check the frozen protocol before the eight locked test seeds were opened. Locked outcomes were not fed back into the design. Each seed generates a 48-hour event path with hourly reviews and daily reviews at hours 24 and 48. The daily review supersedes the hourly one, so each policy faces 48 reviews per seed and the LLM arm makes 384 review decisions in total. Every decision is a separate schema-validated worker call with a recorded session; all 384 sessions are distinct and no fallback response was used.

The four fast variables are the base-fee multiplier, urgency threshold, retry interval, and replacement bump. The two slow variables are the fee cap and target horizon. Fast updates may be requested at any review and re-optimize only the fast group; full updates cover all six variables and are available only at daily reviews. A requested update uses seeded bounded random search over 64 candidates, including the currently deployed parameters, the reported Part~2.2b parameters, and frozen development warm starts. Candidate evaluation uses the trailing three-hour calibration window. A candidate is accepted only if it satisfies the review-time guardrails and improves the operational objective. The guardrails require an inclusion probability of at least the larger of 0.80 and the scenario-specific requirement on each calibration episode, compliance with the scenario fee cap and reported parameter bounds, and zero invalid replacements. If no improving feasible candidate is found, the deployed policy remains active and the failed attempt is recorded.

For episode $t$, let $p_t$ be the realized inclusion probability at the target horizon, $q_t$ the scenario-specific required probability, $f_t$ the paid fee, $c_t$ the fee cap, $o_t$ the overpayment, and $m_t$ the minimum sufficient fee. The frozen operational objective is
\begin{equation}
\label{eq:part23-operational-objective}
\begin{aligned}
J_{\mathrm{op}} &= 0.55L_{\mathrm{rel}}+0.30L_{\mathrm{fee}}+0.15L_{\mathrm{stab}},\\
L_{\mathrm{rel}} &= 0.70\,\overline{\frac{(0.90-p_t)_+}{0.90}}
                  +0.30\,\overline{\mathbb{I}\{p_t<q_t\}},\\
L_{\mathrm{fee}} &= 0.65\,\overline{\frac{f_t}{c_t}}
                  +0.35\,\overline{\frac{o_t}{\max(m_t,1)}},\\
L_{\mathrm{stab}} &= 0.50\,\operatorname{sd}_{\mathrm{pop}}(p_t)
                  +0.50\,\overline{|p_t-p_{t-1}|}.
\end{aligned}
\end{equation}
Here the overline denotes the average over the evaluated episodes, and $\operatorname{sd}_{\mathrm{pop}}$ the population standard deviation across them. The objective contains no optimizer, worker, launch, or update cost. The paired comparisons below use two-sided 95\% Student-$t$ intervals over the eight matched seed-level differences.

\begin{table}[H]
\centering
\small
\setlength{\tabcolsep}{5pt}
\caption{Part 2.3 paired comparisons on the eight locked seeds. Differences are comparator minus LLM-guided selective updating in $J_{\mathrm{op}}$, so positive values favor the LLM policy. Every per-seed difference is positive for all three comparators.}
\label{tab:part23-paired-comparisons}
\begin{tabular}{lrr}
\hline
\textbf{Comparator} & \textbf{Mean difference} & \textbf{95\% CI} \\
\hline
Never update & 0.0050 & [0.0028, 0.0072] \\
Always update & 0.0072 & [0.0039, 0.0106] \\
Rule-based selective & 0.0026 & [0.0009, 0.0042] \\
\hline
\end{tabular}
\end{table}

\begin{table}[H]
\centering
\small
\setlength{\tabcolsep}{5pt}
\caption{Part 2.3 update activity and score components on the locked seeds. Optimizer launches count every re-optimization attempt; accepted updates count launches that changed the deployed parameters. The component values enter $J_{\mathrm{op}}$ through the normalization and weights frozen before the locked run.}
\label{tab:part23-activity}
\begin{tabular}{lrrrrr}
\hline
\textbf{Update policy} & \textbf{Launches/day} & \textbf{Accepted/day} & \textbf{Reliability} & \textbf{Fee} & \textbf{Stability} \\
\hline
Never update & 0.0000 & 0.0000 & 0.0144 & 0.2731 & 0.0504 \\
Rule-based selective & 1.6250 & 1.4375 & 0.0081 & 0.2808 & 0.0421 \\
Always update & 24.0000 & 15.1875 & 0.0218 & 0.2692 & 0.0458 \\
LLM-guided selective & 1.0625 & 1.0625 & 0.0068 & 0.2756 & 0.0399 \\
\hline
\end{tabular}
\end{table}

The activity and component breakdown in Table~\ref{tab:part23-activity} locates each failure mode. Never updating carries the largest stability component and the second-largest reliability component: the frozen parameters lose inclusion reliability once mempool conditions shift persistently, and its minimum inclusion probability (0.7747) and target-horizon satisfaction rate (0.9766) are among the weakest of the four policies. Always updating attempts a refit at all 24 reviews per day, accepts about 15 changes per day, and still records the worst reliability component and the lowest satisfaction rate (0.9479). The LLM policy accepts about one update per day, every launch it requests is accepted, and it attains the best reliability and stability components with a satisfaction rate of 0.9896. Its 384 decisions comprise 367 keep actions, 16 fast updates, and one full update. No update is accepted during a temporary-shock period. Every seed receives at least one targeted update during the persistent-change period, one seed receives a second fast update, and every seed receives one recovery retune after re-entry into the validated band.

The guardrails screen candidate updates on the trailing calibration window; they do not guarantee every subsequent stochastic outcome. On the locked paths, the LLM policy records zero invalid replacements and an aggregate minimum inclusion probability of 0.7994. The latter is a narrow miss of the 0.80 review-time threshold on one seed and is reflected in the reported satisfaction rate.

\paragraph{Connection to the GSMP formalism.}
A generalized semi-Markov scheme specifies states, a finite set of event types, a per-state list of active events, and a transition map; the scheme is driven by clock samples, and it can be read as a deterministic mapping from those samples to event epochs and counts \citep{glasserman1992gsmp}. The Part~2.3 environment instantiates this structure, and Table~\ref{tab:part23-gsmp-mapping} states the correspondence. Two boundaries are deliberate. First, fixing the clock samples fixes the entire event path. This is the formal content of the common-random-number design: the four update policies face the same realized event stream and differ only in which update actions they accept at review epochs. Second, the classical formalism has no decision maker. The supervisor's choice among keep, fast update, and full update at each review is the controlled extension that this experiment adds, which is why the main text writes GSMP-style rather than claiming the classical model.

\begin{table}[H]
\centering
\small
\setlength{\tabcolsep}{5pt}
\caption{Correspondence between the GSMP formalism and the Part 2.3 update environment.}
\label{tab:part23-gsmp-mapping}
\begin{tabular}{p{0.26\textwidth}p{0.66\textwidth}}
\hline
\textbf{GSMP element} & \textbf{Realization in the Part 2.3 environment} \\
\hline
State & Deployed policy parameters and version; rolling base-fee and mempool-pressure estimates; validity-band status; pending transactions; persistence and cooldown counters \\
Event types & Transaction and block events; fee shocks; hidden regime changes; retry and replacement events; hourly and daily reviews \\
Active-event list & What can currently fire: a retry only while a transaction is pending; a fast update only at a review; a full update only at a daily review; the daily review supersedes the hourly one \\
Clock samples & The locked seeds' random draws (arrival timing, shock timing, regime sojourn durations) together with the deterministic review clocks \\
Event epochs and counts & The recorded review trace: per-review observables, accepted updates per day, and optimizer launches \\
Fixed clock samples fix the event path & The common-random-number comparison: identical event streams across all four policies, differing only in the accepted update actions \\
\hline
\end{tabular}
\end{table}

\section{Methodological Details and Executable Instantiation}
\label{app:method-details}
This appendix complements the main methodology by recording the lower-level solvers used in the reported studies and the executable interpretation of the abstract formulation. Appendix~\ref{app:nsga2-details} gives the feasibility-aware NSGA-II procedure used in Part~1 and Parts~2.1--2.2; the Part~2.3 task-specific bounded search is described in Appendix~\ref{app:part23-update-details}. Appendix~\ref{app:method-executable-scope} clarifies how request-conditioned objectives, admissible tuning spaces, validation checks, and archive feedback are implemented without changing the mathematical formulation.

\subsection{Lower-Level NSGA-II Solver}
\label{app:nsga2-details}
For a fixed architecture $d$, the lower-level problem is a constrained multi-objective simulation-optimization problem over the architecture-conditioned domain $\Theta(d)$. In the NSGA-II-based studies, $\Theta(d)$ is a mixed-variable tuning surface: it may contain bounded numeric variables and, when exposed by the specification, discrete or categorical choices. We use NSGA-II with feasibility-aware ranking. For each sampled tuning vector $\theta$, the simulation oracle returns the objective vector $\mathbf{J}_R(d,\theta)$ and guardrail information. Feasible solutions dominate infeasible ones; among infeasible solutions, candidates with lower feasibility penalties are preferred; among feasible solutions, Pareto dominance and crowding distance guide selection. The standard evolutionary loop is summarized in Algorithm~\ref{alg:nsga2}.

\begin{algorithm}[ht]
\caption{Lower-Level Solver: NSGA-II for Fixed Architecture}
\label{alg:nsga2}
\begin{algorithmic}[1]
\REQUIRE Architecture $d$, objective contract $\mathbf{J}_R$, parameter domain $\Theta(d)$, guardrails $\{g_k\}_{k\in\mathcal{K}}$, evaluation budget $B$, population size $N$
\STATE Initialize parent population $\mathcal{Q}_0\subset\Theta(d)$ of size $N$
\STATE Evaluate $\mathbf{J}_R(d,\theta)$ and guardrail violations for all $\theta\in\mathcal{Q}_0$
\STATE Generate offspring $\mathcal{Q}'_0$ by crossover and mutation
\STATE $g\leftarrow0$
\WHILE{evaluations $< B$}
    \STATE $\mathcal{R}_g\leftarrow\mathcal{Q}_g\cup\mathcal{Q}'_g$
    \STATE Evaluate unevaluated offspring by simulation
    \STATE Sort $\mathcal{R}_g$ into non-dominated fronts using constraint-dominance
    \STATE Fill $\mathcal{Q}_{g+1}$ front by front until the next full front would exceed $N$
    \STATE Complete $\mathcal{Q}_{g+1}$ by crowding-distance selection within the boundary front
    \STATE Generate $\mathcal{Q}'_{g+1}$ by crossover and mutation
    \STATE $g\leftarrow g+1$
\ENDWHILE
\RETURN the non-dominated feasible set $\mathcal{P}_d^*$, or the least-violating frontier if no feasible point is found
\end{algorithmic}
\end{algorithm}

\subsection{Executable Instantiation and Scope}
\label{app:method-executable-scope}
The executable system instantiates the abstract formulation through request interpretation, structural validation, simulation evaluation, and archive-backed feedback. A construction request is normalized into an objective contract, admissible tuning surface, and guardrail set before lower-level optimization. Candidate architectures are evaluated only after structural checks; simulation outputs are converted into the request-conditioned objectives used by the optimizer. Engineering artifacts such as schemas, manifests, prompt templates, and reproducibility logs are implementation devices for enforcing Eq.~\eqref{eq:bilevel_problem}; they are not additional mathematical assumptions.

The request-conditioned objective vector $\mathbf{J}_R$ is represented by an executable scoring contract. This contract specifies objective expressions, objective directions, weights, primary progress metrics, and guardrail constraints. Thus, although the main text writes the optimization compactly in minimization form, the executable implementation can represent both minimized and maximized criteria through direction metadata or equivalent loss transformations.

The admissible tuning space $\Theta(d)$ is not treated as a free-form set proposed solely by the LLM. Candidate architectures and tuning surfaces are proposed at the upper level, but lower-level optimization is restricted to variables admitted by the structural specification, parameter metadata, package-level declarations, and validation checks. This distinction preserves the methodological role of the LLM as a proposal mechanism while assigning final admissibility to the executable specification.

The framework does not require one lower-level solver for every downstream task. Part~1 and Parts~2.1--2.2 use mixed-variable NSGA-II with feasibility-aware non-dominated selection, while Part~2.3 uses seeded bounded random search under a fixed per-review budget. In both cases, candidate solutions are evaluated through simulation under the declared objective and guardrail contract. The Part~2.3 instantiation illustrates the solver flexibility stated in Section~\ref{sec:bilevel-formulation}.

The archive used for upper-level feedback is run-scoped. It stores evaluated non-dominated candidates under the active objective contract and selects representative exemplars for the next proposal round. The archive therefore functions as a validated search memory rather than an unrestricted conversational history.

The distinction between construction and downstream use is important for interpreting the experiments. A calibration experiment directly tests digital-twin construction because the target system and evidence define the twin to be built. A downstream policy or strategy experiment may reuse a calibrated twin as a simulator and may still have a bi-level search structure, but it should be interpreted as a DT-enabled optimization task unless the experiment explicitly constructs a new twin under the same structural-validity and objective-contract assumptions.

\end{document}